\documentclass[11pt]{article}

\newif\ifpreprint
\preprinttrue

\ifpreprint
\usepackage[preprint]{acl}
\else
\usepackage[review]{acl}
\fi

\usepackage{times}
\usepackage{latexsym}

\usepackage[T1]{fontenc}
\usepackage[utf8]{inputenc}

\usepackage{microtype}

\usepackage{inconsolata}

\usepackage{multirow}

\usepackage{graphicx}

\usepackage{amsmath}
\usepackage{amsfonts}
\usepackage{float}
\usepackage{booktabs}
\usepackage{hyperref}

\usepackage{amsthm}

\newtheorem{lemma}{Lemma}
\newtheorem{theorem}{Theorem}

\newtheorem{assumption}{Assumption}

\newcommand{\ourmethod}{AttentionPO}

\title{Token-weighted Direct Preference Optimization with Attention}

\author{Chengyu Huang$^1$ \quad
  Zhuohang Li$^2$  \quad
  Sheng-Yen Chou$^1$ \quad
  Claire Cardie$^1$ \\
  $^1$Cornell University
  $^2$Vanderbilt University \\
  \texttt{\{ch2263,sc3379,ctc9\}@cornell.edu}, \texttt{zhuohang.li@vanderbilt.edu} \\}

\begin{document}
\maketitle
\begin{abstract}
Direct Preference Optimization (DPO) aligns Large Language Models with human preferences without the need for a separate reward model. However, DPO treats all tokens in responses equally, neglecting the differing importance of individual tokens. Existing token-level PO methods compute the token weights using either token-position-based heuristic functions or probability estimates given by a separately trained model, which lacks robustness and incurs extra training cost. In contrast, we propose Token-weighted DPO (TwDPO)---a novel training objective grounded on token-weighted RL---and \ourmethod---an instantiation of TwDPO that uses attention from the LLM itself to estimate token weights. \ourmethod~prompts the LLM to serve as a pairwise judge and check where the model attends when comparing the responses. This design makes \ourmethod~\textit{content-aware}, adjusting weights based on response content, and \textit{efficient}, incurring only two extra forward passes per example. Experiment results show that \ourmethod~significantly improves performance on AlpacaEval, MT-Bench, and ArenaHard, surpassing existing Preference Optimization methods.
\ifpreprint
GitHub: \url{https://github.com/HCY123902/AttentionPO}
\fi
\end{abstract}

\section{Introduction}

The alignment of Large Language Models (LLMs) with human preferences and values is critical for ensuring the quality and safety of generated outputs. Early efforts in this domain utilized policy-gradient reinforcement learning (RL) algorithms like Proximal Policy Optimization (PPO) \cite{ppo2017, rlhf2022}. However, these traditional RL methods require training a separate reward model on annotated preference pairs to provide feedback during training \cite{rlhf2022}. To address this complexity, Preference Optimization (PO) methods such as Direct Preference Optimization (DPO) \cite{dpo2023} have emerged. DPO's training objective is derived from the policy gradient, but it learns the reward landscape directly from preference pairs, avoiding the need for an external reward model.

\begin{figure}
    \centering
    \includegraphics[width=\linewidth]{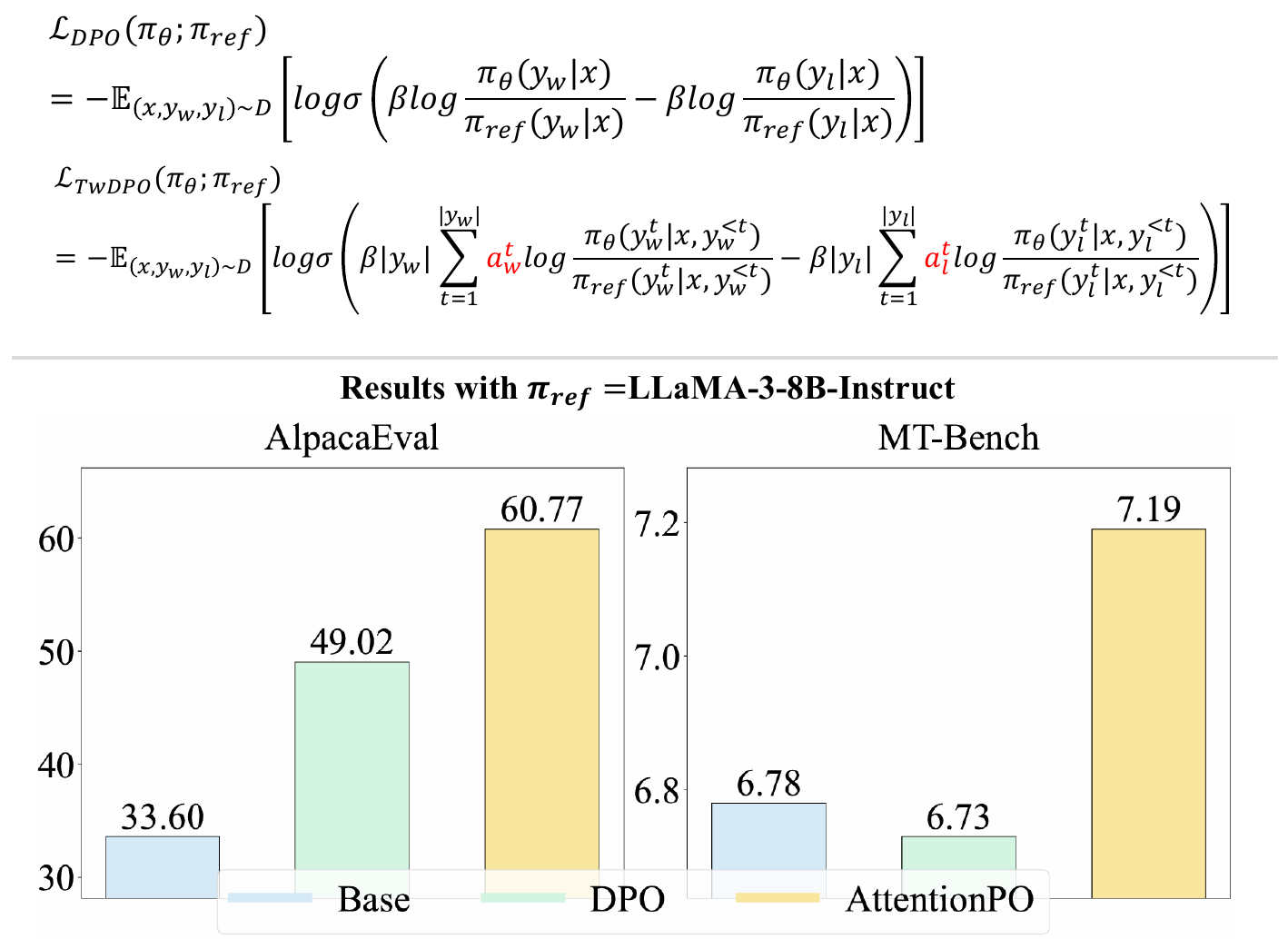}
    \caption{AttentionPO weighs each token by attention, surpassing DPO and various other baselines.}
    \label{fig:placeholder}
    \vspace{-10pt}
\end{figure}

Despite the advancements introduced by DPO and its variants—including IPO \cite{ipo2024}, KTO \cite{kto2024}, and SimPO \cite{simpo2024}—these methods generally treat every token within a response as equally important \cite{simpo2024}. This uniform weighting neglects the fact that different tokens contribute variably to the overall quality of a response \cite{simpo2024}. Consequently, traditional PO methods fail to provide the fine-grained credit assignment necessary to account for the specific importance of individual tokens to human preference.

Recent research has attempted to enable token-level credit assignment through various estimation techniques. Grounded in token-level policy gradient methods, TDPO adds sequence KL terms to the DPO objective \cite{tdpo2024}. Other approaches, such as TIS-DPO \cite{tisdpo2025}, SePO \cite{sepo2025}, and cDPO \cite{cdpo2024}, estimate token importance by contrasting probabilities between different policies or models. Alternatively, TI-DPO utilizes gradient norms and Gaussian priors \cite{tidpo2026}, while D$^2$PO employs heuristic temporal decay functions that prioritize earlier tokens \cite{earlier2025}. However, many of these methods either require the training of additional models to estimate weights \cite{sepo2025, tisdpo2025, cdpo2024} or rely on heuristic functions based on token positions rather than the specific semantic content of the tokens \cite{tidpo2026, earlier2025}.

In this work, we propose Token-weighted DPO (TwDPO), a training objective that is theoretically-grounded on token-weighted RL, and \textbf{\ourmethod}, an instantiation of TwDPO that utilizes the attentions from the LLM itself to estimate token weights. By obtaining attention weights for response tokens from a pairwise judge prompt, \ourmethod~provides a content-aware importance metric. These weights are then normalized and applied during the credit assignment process to better align the model with human preferences.

Experimental results demonstrate the effectiveness of our proposed approach. \ourmethod~achieves significant performance improvements across several models and standard benchmarks. On LLaMA-3-8B-Base-SFT \cite{llama32024, simpo2024}, \ourmethod~improves performance by 12\% (win rate against GPT-4-1106-preview \cite{gpt42023}) on AlpacaEval \cite{alpaca_eval,dubois2024length}, 1.05 (LLM-judged score) on MT-Bench \cite{zheng2023}, and 40\% (win rate against GPT-4-0314) on ArenaHard \cite{arenahard2024, crowdsourced2024}. On LLaMA-3-8B-Insturct, \ourmethod~improves performance by 27\% on AlpacaEval, 1.41  on MT-Bench, and 14\% on ArenaHard.  \ourmethod~also surpasses strong baselines such as SimPO (up to 4\% on AlpacaEval; 0.20 on MT-Bench). \ourmethod~highlights the value of using intrinsic model attentions for precise, token-level preference optimization.

\section{Methodology}
\label{sec:methodology}

\begin{figure*}
    \centering
    \includegraphics[width=\linewidth]{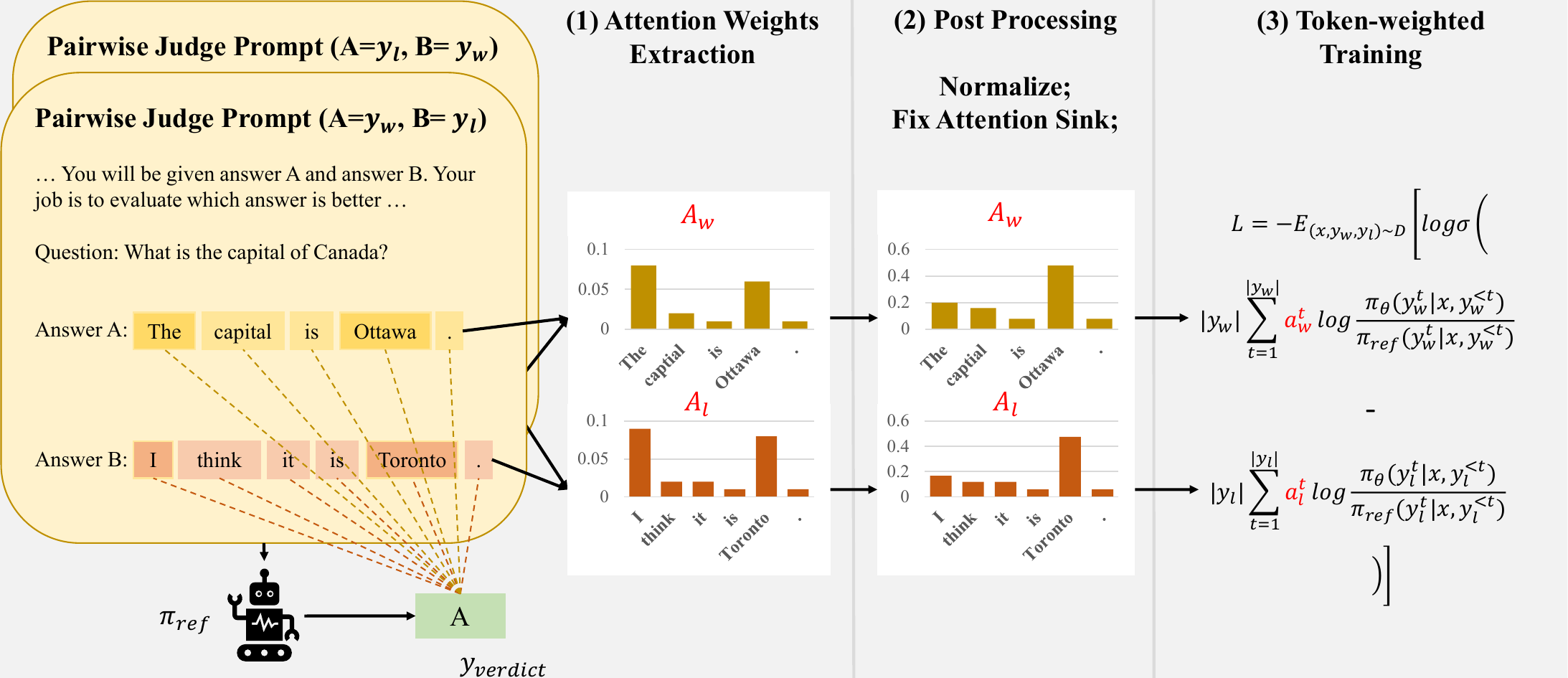}
    \caption{Workflow of \ourmethod. First, we prompt $\pi_{ref}$ to act as a pairwise judge and indicate which response is better in a single output token $y_{verdict}$. We extract $\pi_{ref}$'s attentions at layer $\mathcal{L}$ that attend from $y_{verdict}$ to the response tokens of both $y_w$ and $y_l$. Prompting and weight extraction are repeated two times with the position of $y_w$ and $y_l$ swapped, and the final attention weights $A_w$ and $A_l$ are averaged over the two rounds. Second, we post-process the weights by applying normalization and fixing the attention sink. Finally, we apply TwDPO with the post-processed attention weights.}
    \label{fig:method}
    \vspace{-10pt}
\end{figure*}

We first introduce the notations (\S~\ref{sec:method_notations}) and preliminaries (\S~\ref{sec:method_preliminaries}), then present our TwDPO objective (\S~\ref{sec:method_twdpo}) and its math derivations (\S~\ref{sec:proof}). Finally, we discuss an instantiation of the TwDPO, \ourmethod, which uses attentions as weights. \ourmethod~takes three steps: (1) First, we prompt the initial LLM to act as a pairwise judge to extract the attention weights (\S~\ref{sec:method_awe}); (2) We post-process the attention weights to obtain a weight distribution (\S~\ref{sec:method_pp}); (3) We apply TwDPO using the attention weight distribution. See Figure~\ref{fig:method} for visualization.

\vspace{-5pt}
\subsection{Notations}
\label{sec:method_notations}
\vspace{-5pt}

We initialize our main policy $\pi_{\theta}$ with a reference model $\pi_{ref}$. We train $\pi_{\theta}$ on a preference dataset $\mathcal{D}=\{x^{(i)},y_{w}^{(i)},y_{l}^{(i)}\}_{i=1}^{N}$, where each example contains a prompt $x^{(i)}$, a preferred response $y_{w}^{(i)}$ and a dispreferred response $y_{l}^{(i)}$. Each response $y$ consists of tokens $\{y^{1},\cdots,y^{|y|}\}$.

\subsection{Preliminaries}
\label{sec:method_preliminaries}

Direct Preference Optimization is an offline variant of traditional policy gradient methods to align LLMs with human preferences. It avoids the need for a separate reward model. Instead, it trains LLMs as reward models themselves, on the annotated preference pairs. In particular, DPO maximizes the model's predicted probability of $y_w$ being preferred over $y_l$. Formally, given the preference dataset $\mathcal{D}$, DPO trains $\pi_{\theta}$ using the following objective:
\begin{equation}
    \begin{split}
    \mathcal{L}=-\mathbb{E}_{(x,y_w,y_l)\sim\mathcal{D}}\biggl[log\biggl(p_{\theta}(y_w \succ y_l | x)\biggr)\biggr].
    \end{split}
\end{equation}
where $p_{\theta}(y_w \succ y_l | x)$ is the model's predicted preference probability. Following the Bradley-Terry model \cite{bradley1952},
\begin{equation}
\label{eq:bt}
\small
    \begin{split}
    p_\theta(y_{w}\succ y_{l}|x)&=\frac{\exp(r(x,y_{w}))}{\exp(r(x,y_{w}))+\exp(r(x,y_{l}))}\\
    &=\sigma (r_{\theta}(x, y_w) - r_{\theta}(x, y_l)).
    \end{split}
\end{equation}
where $r_{\theta}(x, y_w)$ is model's predicted reward and $\sigma$ is the sigmoid function. Further, DPO derives the model's predicted reward to be $r_{\theta}(x, y)=\beta log \frac{\pi_{\theta}(y|x)}{\pi_{ref}(y|x)}+\beta log(Z(x))$, where $Z(x)$ is a partition function depending only on $x$. Thus, the final DPO objective is to minimize
\begin{equation}
\label{eq:dpo}
\small
    \begin{split}
    \mathcal{L}=-\mathbb{E}_{(x,y_w,y_l)\sim\mathcal{D}}\biggl[log\sigma\biggl(\beta log \frac{\pi_{\theta}(y_w|x)}{\pi_{ref}(y_w|x)} - \\
    \beta log \frac{\pi_{\theta}(y_l|x)}{\pi_{ref}(y_l|x)}\biggr)\biggr].
    \end{split}
\end{equation}
Equation~\ref{eq:dpo} can be rewritten into the token-level form:
\begin{equation}
\label{eq:dpo_tokens}
\small
    \begin{split}
    \mathcal{L}=-\mathbb{E}_{(x,y_w,y_l)\sim\mathcal{D}}\biggl[log\sigma\biggl(\beta \sum_{t=1}^{|y_w|} \log \frac{\pi_{\theta}(y_w^t|x,y_w^{<t})}{\pi_{ref}(y_w^t|x,y_w^{<t})} - \\
    \beta \sum_{t=1}^{|y_l|} log \frac{\pi_{\theta}(y_l^t|x,y_l^{<t})}{\pi_{ref}(y_l^t|x,y_l^{<t})}\biggr)\biggr].
    \end{split}
\end{equation}
Note that DPO assigns an equal weight of 1 to all the $\frac{\pi_{\theta}(y^t|x,y^{<t})}{\pi_{ref}(y^t|x,y^{<t})}$ terms. This does not consider the varying degrees of importance of different tokens to the final response quality.

\subsection{Token-weighted DPO}
\label{sec:method_twdpo}

Instead, our objective multiplies each probability ratio term in Equation~\ref{eq:dpo_tokens} with a token weight $a^t$:
\begin{equation}
\small
\label{eq:twdpo}
    \begin{split}
    \mathcal{L}=-\mathbb{E}_{\mathcal{D}}\biggl[log\sigma\biggl(&\beta |y_w|\sum_{t=1}^{|y_w|} a_w^{t} \log \frac{\pi_{\theta}(y_w^t|x,y_w^{<t})}{\pi_{ref}(y_w^t|x,y_w^{<t})} \\
    - & \beta |y_l| \sum_{t=1}^{|y_l|} a_l^{t} log \frac{\pi_{\theta}(y_l^t|x,y_l^{<t})}{\pi_{ref}(y_l^t|x,y_l^{<t})}\biggr)\biggr].
    \end{split}
\end{equation}

\subsection{Mathematical Derivation}
\label{sec:proof}

We closely follow the proof of \citet{dpo2023} to derive the token-weighted DPO objective from a token-weighted RL objective. First, we find the closed-form expression of the near-optimal policy $\pi^{*}$ and reward function $r^{*}$ attained by the token-weighted RL objective (\S~\ref{sec:proof_optimum}). Next, we derive the token-weighted DPO objective (\S~\ref{sec:proof_objective}).

\subsubsection{Deriving the Near Optimum of the Token-weighted Objective}
\label{sec:proof_optimum}

DPO \cite{dpo2023} derives the equivalence between their optimal policy and the one attained by a sequence-level RL objective, which maximizes response reward with KL constraints:
\begin{equation}
\label{eq:dpo_rl}
\small
\begin{split}
&\max_{\pi}\mathbb{E}_{x\sim\mathcal{D}, y\sim\pi(\cdot|x)}\left[r(x,y)\right]-\beta \mathbb{D_{KL}}(\pi_{\theta}(y|x) || \pi_{ref}(y|x))\\
&=\max_{\pi}\mathbb{E}_{x\sim\mathcal{D}, y\sim\pi(\cdot|x)}\biggl[r(x,y)-\beta \sum_{t=1}^{|y|}\frac{\pi_{\theta}(y^t|x,y^{<t})}{\pi_{ref}(y^t|x, y^{<t})}\biggr].
\end{split}
\end{equation}

Instead, we show that our objective~\ref{eq:twdpo} gives a model equivalent to the near-optimal policy produced by a token-weighted RL objective. We start by reformulating objective~\ref{eq:dpo_rl}. For a given prompt $x$, a policy $\pi$, and a reference model $\pi_{ref}$, our objective is to find an optimal policy that maximizes the reward under a token-weighted KL penalty. As we will show in \S~\ref{sec:properties}, this objective naturally induces a token-weighted reward function $r$. Our KL penalty weighs the log-probability ratios at each token step $t$ by the weight $a^{t}$:
\begin{equation}
\label{eq:attentionpo_rl}
\small
\begin{split}
&\max_{\pi}\mathbb{E}_{x\sim\mathcal{D},y\sim\pi(\cdot|x)}\biggl[r(x,y) -\\&~~~~~~~~~~\beta|y|\sum_{t=1}^{|y|}a^{t}\log\frac{\pi(y^{t}|x,y^{<t})}{\pi_{ref}(y^{t}|x,y^{<t})}\biggr].
\end{split}
\end{equation}
Note that $a^t$ is a proper distribution and sums up to 1. In light of this, we scale the KL term by the sequence length $|y|$ so that the total weight of the logprob ratio terms $|y|\sum_{t=1}^{|y|}a^t$ equals $|y|$, which is consistent with the DPO objective~\ref{eq:dpo_rl}.

To solve this mathematically, we define a sequence-level "weighted" log-probability\footnote{$\tilde{\pi}(y|x)$ and $\tilde{\pi}(y|x)_{ref}$ do not need to be proper distribution, but can be normalized to be one. See Appendix~\ref{app:weighted_prob_term}.} for both the policy and the reference model:
\begin{equation}
\small
\begin{split}
\log\tilde{\pi}(y|x) &= |y|\sum_{t=1}^{|y|}a^{t}\log\pi(y^{t}|x,y^{<t}), \\
\log\tilde{\pi}_{ref}(y|x) &= |y|\sum_{t=1}^{|y|}a^{t}\log\pi_{ref}(y^{t}|x,y^{<t}).
\end{split}
\end{equation}
Substituting these into our objective, we can follow the standard derivation steps:
\begin{equation}
\small
\begin{split}
&\max_{\pi}\mathbb{E}_{x\sim\mathcal{D}}\mathbb{E}_{y\sim\pi(\cdot|x)}\left[r(x,y)-\beta\log\frac{\tilde{\pi}(y|x)}{\tilde{\pi}_{ref}(y|x)}\right] \nonumber \\
&= \min_{\pi}\mathbb{E}_{x\sim\mathcal{D}}\mathbb{E}_{y\sim\pi(\cdot|x)}\left[\log\frac{\tilde{\pi}(y|x)}{\tilde{\pi}_{ref}(y|x)}-\frac{1}{\beta}r(x,y)\right] \nonumber \\
&= \min_{\pi}\mathbb{E}_{x\sim\mathcal{D}}\mathbb{E}_{y\sim\pi(\cdot|x)}\biggl[\log\frac{\tilde{\pi}(y|x)}{\frac{1}{Z(x)}\tilde{\pi}_{ref}(y|x)\exp(\frac{1}{\beta}r(x,y))}\\&~~~~~~~~~~~~~~~~~~~~~~~~~~~~~~~~~~~-\log Z(x)\biggr],
\end{split}
\end{equation}
where $Z(x)=\sum_{y}\tilde{\pi}_{ref}(y|x)\exp(\frac{1}{\beta}r(x,y))$ is the partition function. 

The near optimal weighted policy $\tilde{\pi}^{*}$ (we analyze the error bounds between $\tilde{\pi}^{*}$ and the intractable true optimal policy in Appendix~\ref{sec:error_bound}) that minimizes this expression satisfies:
\begin{equation}
\label{eq:weighted_pi}
\tilde{\pi}^{*}(y|x)=\frac{1}{Z(x)}\tilde{\pi}_{ref}(y|x)\exp\left(\frac{1}{\beta}r(x,y)\right).
\end{equation}
By taking the logarithm of both sides and rearranging the terms, we can express the true reward $r^{*}(x,y)$ entirely in terms of the optimal policy:
\begin{equation}
\label{eq:attention_po_reward_abbrev}
r^{*}(x,y)=\beta\log\frac{\tilde{\pi}^{*}(y|x)}{\tilde{\pi}_{ref}(y|x)}+\beta\log Z(x).
\end{equation}
Substituting the expanded definition of our weighted log-probabilities back into this reward expression yields:
\begin{equation}
\label{eq:attention_po_reward}
\small
r^{*}(x,y)=\beta|y|\sum_{t=1}^{|y|}a^{t}\log\frac{\pi^{*}(y^{t}|x,y^{<t})}{\pi_{ref}(y^{t}|x,y^{<t})}+\beta\log Z(x).
\end{equation}

\subsubsection{Deriving the Objective Under the Bradley-Terry Model}
\label{sec:proof_objective}

We substitute our token-weighted reparameterization of $r^{*}(x,y)$ into the Bradley-Terry preference model (Equation~\ref{eq:bt}):
\begin{equation}
\small
    \begin{split}
    & p^{*}(y_{w}\succ y_{l}|x) = \sigma (r^{*}(x, y_w) - r^{*}{\theta}(x, y_l))\\
    &=\sigma\Bigg(\left(\beta|y_{w}|\sum_{t=1}^{|y_{w}|}a_{w}^{t}\log\frac{\pi^{*}(y_{w}^{t}|x,y_{w}^{<t})}{\pi_{ref}(y_{w}^{t}|x,y_{w}^{<t})}+\beta\log Z(x)\right) \nonumber \\
    &\quad - \left(\beta|y_{l}|\sum_{t=1}^{|y_{l}|}a_{l}^{t}\log\frac{\pi^{*}(y_{l}^{t}|x,y_{l}^{<t})}{\pi_{ref}(y_{l}^{t}|x,y_{l}^{<t})}+\beta\log Z(x)\right)\Bigg).
    \end{split}
\end{equation}
Crucially, the partition function terms $\beta\log Z(x)$ cancel out, leaving:
\begin{equation}
\label{eq:reward_prob}
\small
\begin{split}
p^{*}(y_{w}\succ y_{l}|x)=\sigma\biggl(\beta|y_{w}|\sum_{t=1}^{|y_{w}|}a_{w}^{t}\log\frac{\pi^{*}(y_{w}^{t}|x,y_{w}^{<t})}{\pi_{ref}(y_{w}^{t}|x,y_{w}^{<t})}\\
-\beta|y_{l}|\sum_{t=1}^{|y_{l}|}a_{l}^{t}\log\frac{\pi^{*}(y_{l}^{t}|x,y_{l}^{<t})}{\pi_{ref}(y_{l}^{t}|x,y_{l}^{<t})}\biggr).
\end{split}
\end{equation}
To frame this as a maximum likelihood objective for training a parameterized policy $\pi_{\theta}$, we take the negative log-likelihood over the preference dataset $\mathcal{D}$. This leads to our final objective in Equation~\ref{eq:twdpo}.

% \begin{equation}
% \mathcal{L}=-\mathbb{E}_{(x,y_{w},y_{l})\sim\mathcal{D}}\left[\log\sigma\left(\beta|y_{w}|\sum_{t=1}^{|y_{w}|}a_{w}^{t}\log\frac{\pi_{\theta}(y_{w}^{t}|x,y_{w}^{<t})}{\pi_{ref}(y_{w}^{t}|x,y_{w}^{<t})}-\beta|y_{l}|\sum_{t=1}^{|y_{l}|}a_{l}^{t}\log\frac{\pi_{\theta}(y_{l}^{t}|x,y_{l}^{<t})}{\pi_{ref}(y_{l}^{t}|x,y_{l}^{<t})}\right)\right]
% \end{equation}

\subsubsection{Properties}
\label{sec:properties}

\paragraph{Reward function is token-weighted.} Equation~\ref{eq:attention_po_reward_abbrev} shows that $r$ weighs tokens by $a^t$. For tokens with higher $a^t$, their logprob ratio between $\pi^*(y^t|x,y^{<t})$ and $\pi_{ref}(y^t|x,y^{<t})$ significantly affect $r$, while tokens with lower $a^t$ affect $r$ less.

\paragraph{Token-weighted KL penalty enables locally-adaptive trust region.}
% Our KL penalty in objective~\ref{eq:attentionpo_rl} weighs the KL divergence at the token level ($a^{t}$). The model is allowed to deviate more significantly from the reference policy on tokens with low attention weights (low importance) and is more strictly constrained on high-attention tokens (high importance). Such adaptive weighting is necessary because important tokens dominate the reward function and are more prone to over-optimization or reward hacking. Intuitively, the token-weighted KL term balances the effect of the token-weighted reward function. For tokens with high weights, there are more incentives (rewards) to deviate from the reference distribution, but at the same time, more penalties for drifting too far.
Our KL penalty in objective~\ref{eq:attentionpo_rl} scales the KL divergence at the token level via $a^{t}$. Consequently, the model is granted greater freedom to deviate from the reference policy on low-importance tokens, while being strictly constrained on high-importance ones. While restricting the most critical tokens may seem counterintuitive, this adaptive weighting is essential because important tokens dominate the implicitly derived reward function, and thus are highly susceptible to reward hacking and over-optimization. The token-weighted KL term acts as a locally adaptive regulator: for high-weight tokens, the massive incentives to deviate are safely balanced by proportionally stricter penalties, ensuring the model only updates when the expected reward justifies the risk.

The above suggests that objective~\ref{eq:attentionpo_rl} is indeed a token-weighted RL objective in terms of both the reward function and KL penalty. Consequently, both its derived policy and the near-optimal policy derived from objective~\ref{eq:twdpo} learn more signals from important tokens, which is what we desire. We further provide an analysis on the gradient in Appendix~\ref{app:gradient_analysis}. While we use attention as the weights, there are no assumptions about how the weights are derived, and \ourmethod~is only one possible instantiation.

% \subsection{Attention Weights Extraction}
\subsection{Token Weights Extraction Through Attention Scores}
\label{sec:method_awe}

We hypothesize that if we ask the LLM to judge the response quality, the tokens that get more attention are more important. As such, we prompt $\pi_{ref}$ with a general pairwise judge prompt (See Appendix~\ref{fig:pairwise_judge_prompt}) that asks it to predict which of $y_w$ and $y_l$ has better quality. We explicitly require $\pi_{ref}$ to output the identifier ("A" or "B") of the better response, and we denote this single token as $y_{verdict}$. To avoid bias, we hide the preference labels in the prompt and call $\pi_{ref}$ two rounds, each round with the positions of $y_w$ and $y_l$ swapped. For each round, we do a single forward pass and obtain the attention weights at a fixed layer $L$. We extract the attention weights from $y_{verdict}$ to all the tokens in $y_w$ and denote them as $A_w^{round\_r}=[a_w^{1},\cdots,a_w^{|y_w|}]$. Similarly, we extract the attention weights from $y_{verdict}$ to all the tokens in $y_l$ as $A_l^{round\_r}=[a_l^{1},\cdots,a_l^{|y_l|}]$. For simplicity, we compute the final attention weight of each token as the mean attention weight for that token across attention heads.

For rounds $r\in\{1,2\}$ ,This produces $A_w^{round\_1}$, $A_w^{round\_2}$, $A_l^{round\_1}$, $A_l^{round\_2}$. The final attention is the average of the two rounds:
\begin{equation}
    \begin{split}
        a_w^t &= \frac{1
}{2}a_w^{t,round\_1}+\frac{1}{2}a_w^{t,round\_2}, \\
        a_l^t &= \frac{1
}{2}a_l^{t,round\_1}+\frac{1}{2}a_l^{t,round\_2}.
    \end{split}
\end{equation}

\subsection{Post-Processing}
\label{sec:method_pp}

\paragraph{Normalization.} $\pi_{ref}$ also attends to other tokens in the pairwise judge prompt and so neither $A_w$ nor $A_l$ form a proper distribution (i.e., it does not sum up to 1.0). Therefore, we normalize the attention weights $A_w$ and $A_l$: $a_w^{t}\leftarrow a_w^{t}/(\sum_{q=1}^{|y_w|} a_w^q)$, $a_l^{t}\leftarrow a_l^{t}/(\sum_{q=1}^{|y_l|} a_l^q)$.

\paragraph{Fixing Attention Sink.} Attention sink is a phenomenon where the LLMs predominantly attend to the initial tokens \cite{xiao2024}. Manual inspection on our dataset $\mathcal{D}$ suggests that these initial tokens are usually of low importance, consisting of starting phrases. As such, for any response with at least $K'$ tokens, we reset the attention weights of the first $K$ tokens of each response $y$ to the average weight $1/|y|$ and renormalize the remaining tokens: 
\begin{equation}
    \small
    \begin{split}
    \forall t\in \{K+1, \cdots, |y_w|\}, a_w^{t} & \leftarrow a_w^{t} \frac{1-\frac{K}{|y_w|}}{\sum_{q=K+1}^{|y_w|} a_w^q} \\
    \forall t\in \{K+1, \cdots, |y_l|\}, a_l^{q} & \leftarrow a_l^{q} \frac{1-\frac{K}{|y_l|}}{\sum_{q=K+1}^{|y_l|} a_l^q}
    \end{split}
\end{equation}
% \paragraph{Token Matching.} During training, due to changes in the tokenization context, certain tokens may not be matched to an attention score. For these tokens, we assign a weight of 0. See Appendix~\ref{app:token_matching} for details.

\section{Experiment Setup}
\label{sec:setup}

\paragraph{Models and Datasets.} We follow the settings from \citet{simpo2024}. For the initial model $\pi_{ref}$, we experiment with 
% zephyr-7b-sft-full \cite{tunstall2023}, which is a model supervised-finetuned from Mistral-7B-v0.1 \cite{mistral2023} on UltraChat-200K \cite{ultrachat2023}; Mistral-7B-Instruct-v0.2 \cite{mistral2023}; 
\textbf{Llama-3-8B-Base-SFT} from \citet{simpo2024} (princeton-nlp/Llama-3-Base-8B-SFT), which is supervised-finetuned from the pretrained model LLaMA-3-8B \cite{llama32024} on UltraChat-200K \cite{ultrachat2023}; and \textbf{LLaMA-3-8B-Instruct} \cite{llama32024} (meta-llama/Meta-Llama-3-8B-Instruct).

For the SFT model (Llama-3-8B-Base-SFT), we apply \ourmethod~on the UltraFeedback binarized dataset \cite{ultrafeedback2024} (HuggingFaceH4/ultrafeedback\_binarized). This dataset contains 61,135/2,000 training/validation examples covering diverse instruction-following tasks such as creative writing, document-assisted writing, and open-ended QA. Each query is accompanied by four responses generated by diverse LLMs and graded by GPT-4 in terms of instruction-following, helpfulness, honesty, and truthfulness. $y_w$ is set to the response with the highest overall score, and $y_l$ is chosen randomly from the remaining three.

For the Instruction-Tuned model (LLaMA-3-8B-Instruct), we use the dataset produced by \citet{simpo2024} (princeton-nlp/llama3-ultrafeedback). It contains 59,876/1,961 training/validation examples. The questions are subsampled from the binarized datasets, but for each question, five responses are sampled from $\pi_{ref}$ and then scored by a reward model PairRM \cite{llmblender2023}. $y_w$ is set to the response with the highest score and $y_l$ is set to the one with the lowest score.

\begin{table*}[]
\small
    \centering
    \begin{tabular}{lccccccc}
        \toprule
         & \multicolumn{3}{c}{$\pi_{ref}$=LLaMA-3-8B-Base-SFT} & \multicolumn{3}{c}{$\pi_{ref}$=LLaMA-3-8B-Instruct} \\
         \cmidrule(lr){2-4} \cmidrule(lr){5-7}
         & Std & Max & Len & Std & Max & Len \\
         \midrule
        $y_w$ & 0.0177 & 0.1242 & 269.0895 & 0.0126 & 0.0915 & 356.9848 \\
        $y_l$ & 0.0175 & 0.1202 & 235.3978 & 0.0137 & 0.0959 & 365.8176 \\
        \bottomrule
    \end{tabular}
    \caption{Statistics of the attention weights after post-processing in \S~\ref{sec:method_pp}. Std: Standard deviation of token weights within a response. Max: Maximum weights; Len: Response length; Metrics are averaged across the training set.}
    \label{tab:aw_stat}
\end{table*}

\begin{table*}[]
\small
    \centering
    \begin{tabular}{lclclclc}
        \toprule
         \multicolumn{4}{c}{$\pi_{ref}$=LLaMA-3-8B-Base-SFT} & \multicolumn{4}{c}{$\pi_{ref}$=LLaMA-3-8B-Instruct} \\
         \cmidrule(lr){1-4} \cmidrule(lr){5-8}
         \multicolumn{2}{c}{$y_w$} & \multicolumn{2}{c}{$y_l$} & \multicolumn{2}{c}{$y_w$} &
         \multicolumn{2}{c}{$y_l$} \\
         \midrule
         Token & Weight & Token & Weight & Token & Weight & Token & Weight \\
         \midrule
         True & 0.1237 & Ye & 0.8212 & (Note & 0.1269 & (Note & 0.1276 \\
         idence & 0.1227 & Neutral & 0.1213 & YES & 0.0918 & Yes & 0.0853 \\
         Neutral & 0.1184 & ye & 0.1065 & "What & 0.0844 & "What & 0.0673 \\
         POS & 0.1148 & Completion & 0.0987 & doesn & 0.0643 & doesn & 0.0618 \\
         False & 0.1016 & idence & 0.0886 & wouldn & 0.0606 & No & 0.0598 \\
         Completion & 0.0997 & No & 0.0845 & False & 0.0572 & !ĊĊ & 0.0529 \\
         No & 0.0990 & True & 0.0796 & Yes & 0.0563 & didn & 0.0528 \\
         Negative & 0.0788  & Ent & 0.0721 & No & 0.0556 & wouldn & 0.0510 \\
         Positive & 0.0668 & Negative & 0.0692 & Explanation & 0.0505 & don & 0.0497 \\
         Yes & 0.0591 & False & 0.0686 & don & 0.0492 & "The & 0.0447 \\
        \bottomrule
    \end{tabular}
    \caption{Most weighted tokens after post-processing in \S~\ref{sec:method_pp}. Weight: Weights averaged across all occurrences of the token in the training set. We only keep tokens that occur at least 100 times.}
    \label{tab:aw_mt}
    \ifpreprint
        \vspace{-10pt}
    \else
    \fi
\end{table*}

\paragraph{Attention Weights.} We set $L$ to the index of the last layer for every model (i.e., we take the attention weights from the last layer). To fix the attention sink issue, we use $K=1$ and $K'=5$. That is, for response $y$ that has at least 5 tokens, we set the attention weight of its first token to $1/|y|$. The rationale is that for our models, the attention sink primarily concentrates on the first token. For LLaMA3-8B-Base-SFT, significant attention weights are on the first token (12.55\%), whereas the weights on the second (3.35\%) and third (2.55\%) tokens are much smaller. For LLaMA-3-8B-Instruct, the proportion is 8.25\%/2.11\%/1.55\% for the first, second, and third tokens, respectively.

We show detailed statistics of the attention weights in Table~\ref{tab:aw_stat}. The standard deviation of the weights from both SFT and Instruct models is above 0.01, suggesting that the weights vary across tokens. We also show the ten most weighted tokens in Table~\ref{tab:aw_mt}. For the SFT model, 7/10 tokens are related to binary judgment words (e.g., "Yes", "True") for both $y_w$ and $y_l$. The highly weighted tokens of the Instruct model not only include these but also negation words (e.g., "doesn").

\paragraph{Training Details.} We set learning rate to $1e-6$ with a cosine scheduler and a warmup ratio of 0.1.  We use the AdamW optimizer \cite{adamw2019}. The KL regularizer $\beta$ is set to $0.005$. See more details in Appendix~\ref{app:hyperparameters}.

\paragraph{Baselines.} We compare \ourmethod~against the initial model $\pi_{ref}$ and models trained with common PO methods, including \textbf{RRHF} \cite{rrhf2023}, \textbf{SLiC-HF} \cite{slichf2023}, \textbf{DPO} \cite{dpo2023}, \textbf{TDPO} \cite{tdpo2024}, \textbf{TI-DPO} \cite{tidpo2026}, \textbf{OTPO} \cite{otpo2025}, \textbf{IPO} \cite{ipo2024}, \textbf{CPO} \cite{cpo2024}, \textbf{KTO} \cite{kto2024}, \textbf{ORPO} \cite{orpo2024}, \textbf{R-DPO} \cite{rdpo2024}, and \textbf{SimPO} \cite{simpo2024}. Except TDPO, TI-DPO, and OTPO, we adopt the training hyperparameters from \citet{simpo2024}, which have been separately searched to be optimal.

\paragraph{Evaluation.} Following prior work \cite{simpo2024, sepo2025, huang2025}, we evaluate the models on AlpacaEval \cite{alpaca_eval}, MT-Bench \cite{zheng2023}, and ArenaHard \cite{arenahard2024}. For AlpacaEval, we report the raw and length-controlled win rate of our model against GPT-4-1106-preview. For MT-Bench, we report the LLM-judged quality score of our model's generated response on a scale of 1 to 10. For ArenaHard, we report the win rate of our model against GPT4-0314. We use GPT-4o-mini \cite{gpt4o2024} as the judge model for these benchmarks. See more evaluation details in Appendix~\ref{app:hyperparameters}.

% WPO: wzhouad/zephyr-7B-WPO-FP

\ifpreprint
\vspace{-5pt}
\else
\fi
\section{Results}
\ifpreprint
\vspace{-5pt}
\else
\fi

\label{sec:results}

We first present the main results (\S~\ref{sec:main_results}), followed by comparing our method with other choices of token weights (\S~\ref{sec:varying_tw}), and finally conducting an ablation study (\S~\ref{sec:ablations}).

\subsection{Main Results}
\label{sec:main_results}

\begin{table*}[]
    \small
    \centering
    \resizebox{\linewidth}{!}{
    \begin{tabular}{lcccccccccc}
        \toprule
         & \multicolumn{5}{c}{$\pi_{ref}$=LLaMA-3-8B-Base-SFT} & \multicolumn{5}{c}{$\pi_{ref}$=LLaMA-3-8B-Instruct} \\
         \cmidrule(lr){2-6} \cmidrule(lr){7-11}
         & \multicolumn{2}{c}{AlpacaEval} & MT & ArenaHard & \multirow{2}{*}{\textbf{Avg Rank $\downarrow$}} & \multicolumn{2}{c}{AlpacaEval} & MT & ArenaHard & \multirow{2}{*}{\textbf{Avg Rank $\downarrow$}} \\
         \cmidrule(lr){2-3} \cmidrule(lr){4-4} \cmidrule(lr){5-5} \cmidrule(lr){7-8} \cmidrule(lr){9-9} \cmidrule(lr){10-10}
         & LR $\uparrow$ & WR $\uparrow$ & Score $\uparrow$ & WR $\uparrow$ & & LR $\uparrow$ & WR $\uparrow$ & Score $\uparrow$ & WR $\uparrow$ & \\
         \midrule
         Base       & 9.20 & 4.63 & 5.47 & 9.92 & 13.75 & 33.60 & 32.60 & 6.78 & 37.57 & 12.25 \\
         +RRHF      & 11.29 & 6.63 & 6.18 & 26.00 & 9.25 & 43.75 & 37.65 & 6.63 & 41.91 & 10.00 \\ 
         +SLiC-HF   & 16.73 & 12.08 & 5.95 & 13.08 & 10.00 & 38.38 & 37.03 & 6.76 & 42.76 & 10.13 \\
         +DPO       & 22.88 & 17.70 & \textbf{6.54} & 38.54 & \underline{3.25} & 49.02 & 44.72 & 6.73 & 53.98 & 6.00 \\
         +TDPO & 11.96 & 6.18 & 5.68 & 12.09 & 11.75 & 37.74 & 37.34 & 6.64 & 41.00 & 11.50 \\
         +TI-DPO & 9.94 & 4.77 & 5.29 & 10.14 & 13.25 & 35.41 & 34.57 & 6.80 & 38.73 & 10.75 \\
         +OTPO & \underline{24.24} & 14.43 & 6.22 & 46.31 & 4.50 & 51.26 & 44.65 & 6.91 & 52.60 & 4.50 \\
         +IPO       & 20.69 & 19.73 & 6.23 & 34.56 & 4.75 & 48.94 & 47.47 & \underline{6.99} & 50.50 & 4.50 \\
         +CPO       & 18.94 & \underline{24.56} & 5.38 & 36.04 & 4.88 & 41.26 & 45.13 & 6.67 & 52.07 & 7.50 \\
         +KTO       & 18.24 & 15.05 & 5.93 & 18.28 & 8.75 & 43.10 & 39.73 & 6.79 & 48.98 & 7.38 \\
         +ORPO      & 19.36 & 13.47 & 5.72 & 17.32 & 9.25 & 40.92 & 36.95 & 6.76 & 44.64 & 9.88 \\
         +R-DPO     & 20.94 & 15.57 & 6.33 & 28.72 & 5.25 & 52.11 & 46.98 & 6.79 & \textbf{54.98} & \underline{3.38} \\
         +SimPO     & \textbf{32.26} & \textbf{29.11} & 6.13 & \underline{47.01} & 3.63 & \underline{56.79} & \underline{47.73} & 6.51 & \underline{54.94} & 5.00 \\
         \textbf{+\ourmethod} & 21.55 & 20.23 & \underline{6.52} & \textbf{49.72} & \textbf{2.25} & \textbf{60.77} & \textbf{58.29} & \textbf{7.19} & 52.06 & \textbf{2.25} \\
         \bottomrule
    \end{tabular}
    }
    \caption{Main results. Base: $\pi_{ref}$; MT: MT-Bench; AlpacaEval: LR is the length-controlled win rate and WR is the raw win rate against GPT-4-1106-Preview; MT-Bench: LLM-judged quality score of the model's responses, on a scale of 1 to 10; ArenaHard: Win rate against GPT-4-0314.  Avg Rank: arithmetic mean of the method's ranks across all individual metrics. AttentionPO achieves the best overall ranking across both Base-SFT and Instruct settings. See Table~\ref{tab:main_results_multiple_runs} for results with multiple evaluation runs.}
    \label{tab:main_results}
    \vspace{-10pt}
\end{table*}

As shown in Table~\ref{tab:main_results}, \textbf{\ourmethod~significantly improves the performance over the base model $\pi_{ref}$ in all settings}. On LLaMA-3-8B-Base-SFT, performance increases by 12\% on AlpacaEval, 1.05 on MT-Bench, 40\% on ArenaHard. On LLaMA-3-8B-Instruct, performance increases by 27\% on AlpacaEval, 0.41 on MT-Bench, and 14\% on ArenaHard. Furthermore, \textbf{\ourmethod~outperforms the baselines in most settings}. For the SFT model, \ourmethod~ranks first on ArenaHard, second on MT-Bench, and third on AlpacaEval. For the Instruct model, \ourmethod~outperforms all baselines on AlpacaEval and MT-Bench. Across the benchmarks, \ourmethod~achieves the best overall rank for both models.

% \subsection{Varying the source of token weights}
\subsection{Impact of Token Weight Sources}
\label{sec:varying_tw}
There are alternative ways to obtain token-level weights. Prior work \cite{cdpo2024, sepo2025, tisdpo2025, earlier2025, tidpo2026} either requires training a separate model to estimate weights or relies on strong heuristics. Instead, we stick to our setting where the weights are attained through self-judging, but we experiment with various changes to the current design. This includes (1) Extracting the attention weights from a different hidden layer $L$ of $\pi_{ref}$; (2) Using attention rollout \cite{attentionrollout2020} to aggregate the weights across different hidden layers; (3) Prompting $\pi_{ref}$ to verbally assign a weight for each token. For these experiments, we use LLaMA-3-8B-Instruct as an example.

\begin{figure}
    \centering
    \includegraphics[width=\linewidth]{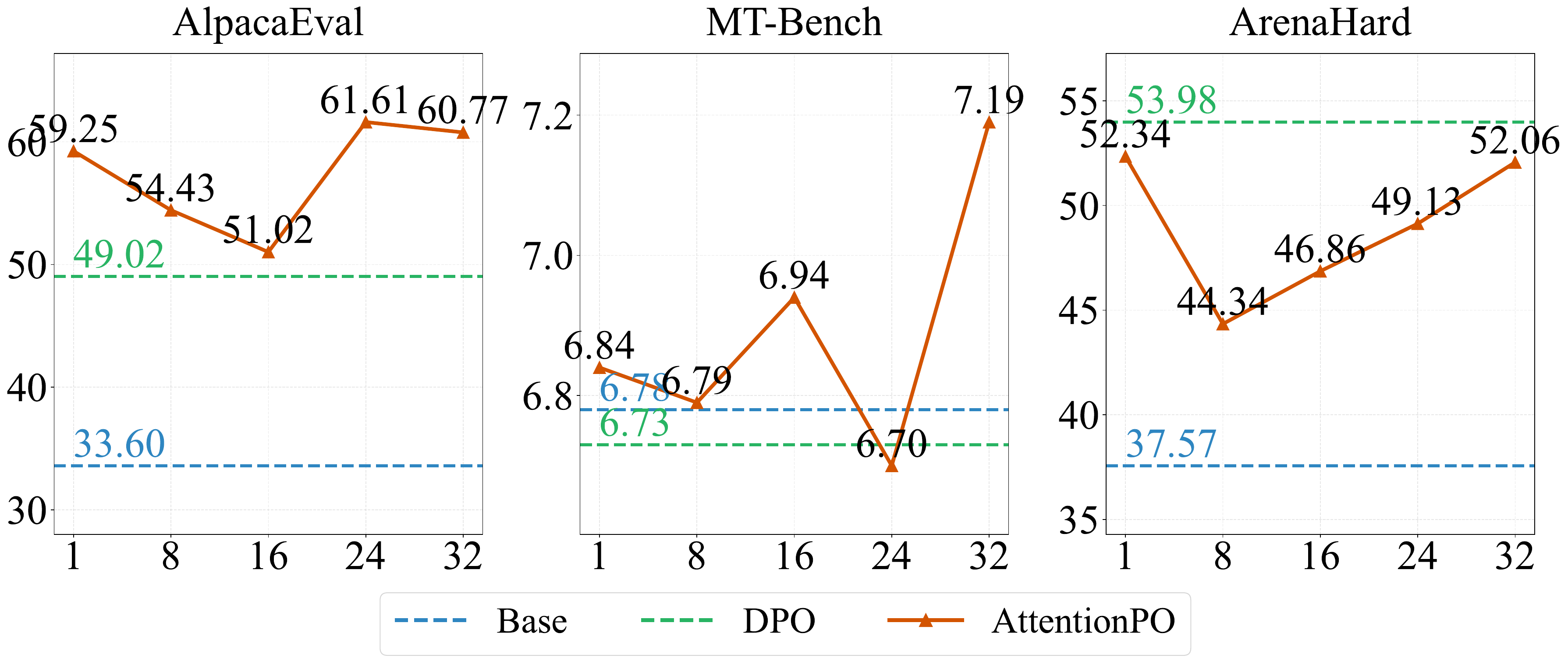}
    \caption{Results for different layers. x-axis: index of the layer from which attention weights are extracted; y-axis: benchmark performance.}
    \label{fig:res_cl}
    \vspace{-10pt}
\end{figure}

\paragraph{Changing layers.} LLaMA-3-8B-Instruct consists of 32 hidden layers. We use $L=32$ for our main experiments. Here, we experiment with $L \in \{1,8,16,24,32\}$ and present the results in Figure~\ref{fig:res_cl}. We find that \textbf{the first and last layers perform better than the middle ones, with the last layer achieving the best results}. We leave more investigation on this to future work.

\paragraph{Attention rollout (\texttt{w/ attn rollout}).} Attention rollout aggregates the attentions across all the layers to estimate the total attentions from tokens in the final output layer to tokens in the initial input layer. Table~\ref{tab:res_ws} shows that using their aggregated weights underperforms our method. This is potentially associated with our previous observation that middle-layer attentions are not suitable for \ourmethod, since attention rollout incorporates attentions from these layers into the final weights.

\paragraph{Verbalized self-judged weights (\texttt{w/ verbal w}).} Instead of using attention, we ask $\pi_{ref}$ to verbally assign an importance score on a scale of 1 to 5 to each token in the response. We observe that it is difficult for the model to precisely give a score for every token (e.g., malformed scoring report; not being able to recall every token in the response). Therefore, we adapt a coarser granularity and ask the model to first partition the response into parts and then assign a score for each part (See prompts in Appendix~\ref{app:prompt}). Even with a coarser granularity, $\pi_{ref}$ still cannot perfectly recall the response tokens. We filter out examples where fewer than 90\% of the tokens are matched with a verbalized weight for either $y_w$ or $y_l$. After filtering, we get 10,111/343 train/validation examples. To compensate for a smaller dataset, we increase the number of training epochs from 1 to 5 and validate at each epoch.
 \begin{table}[]
    \small
    \centering
    \resizebox{\linewidth}{!}{
    \begin{tabular}{lcccc}
        \toprule
         & \multicolumn{4}{c}{$\pi_{ref}$=LLaMA-3-8B-Instruct}\\
         & \multicolumn{2}{c}{AlpacaEval} & MT & ArenaHard \\
         \cmidrule(lr){2-3} \cmidrule(lr){4-4} \cmidrule(lr){5-5} 
         & LR $\uparrow$ & WR $\uparrow$ & Score $\uparrow$ & WR $\uparrow$ \\
         \midrule
         Base       & 33.60 & 32.60 & 6.78 & 37.57 \\
         +DPO & 49.02 & 44.72 & 6.73 & \underline{53.98} \\
         +\ourmethod & \textbf{60.77} & \textbf{58.29} & \textbf{7.19} & 52.06 \\
         ~~w/ attn rollout & 54.84 & 54.87 & 6.71 & 48.23 \\
         ~~w/ verbal w & \underline{57.95} & \underline{57.76} & \underline{6.89} & \textbf{61.41} \\
         \bottomrule
    \end{tabular}}
    \caption{Results with different weight sources.}
    \label{tab:res_ws}
    \vspace{-5pt}
\end{table}
Table~\ref{tab:res_ws} shows that this method outperforms ours on ArenaHard but underperforms on AlpacaEval and MT-Bench. A potential reason is the use of a coarser granularity. We leave more investigation on the viability of this approach to future work.

\subsection{Ablation Study}
\label{sec:ablations}
\paragraph{Restoring the attention sink (\texttt{w/ attn sink}).} We restore the attention sink, in which case the first token occupies roughly 8.25\% of the total weights. Table~\ref{tab:res_ablations} shows that this variant underperforms our method on all benchmarks, indicating the benefits of averaging out the attention sink.

\paragraph{Adding length normalization (\texttt{w/ len norm}).} Prior work \cite{sepo2025, simpo2024} argues that DPO is biased toward lengthy responses (i.e., lengthy responses receive more gradient updates than concise ones). They add a length normalizer to the objective. Similarly, we remove the $|y|$ multiplier in front of the logprob ratio term for $y_w$ and $y_l$, giving a weighted, length-normalized variant:
\begin{equation}
\label{eq:twdpo_len_norm}
\small
    \begin{split}
    \mathcal{L'}=-\mathbb{E}_{\mathcal{D}}\biggl[log\sigma\biggl(&\beta \sum_{t=1}^{|y_w|} a_w^{t} \log \frac{\pi_{\theta}(y_w^t|x,y_w^{<t})}{\pi_{ref}(y_w^t|x,y_w^{<t})} \\
    - & \beta \sum_{t=1}^{|y_l|} a_l^{t} log \frac{\pi_{\theta}(y_l^t|x,y_l^{<t})}{\pi_{ref}(y_l^t|x,y_l^{<t})}\biggr)\biggr]
    \end{split}
\end{equation}
Adding the length normalization effectively reduces the KL regularization constant $\beta$. Therefore, we increase $\beta$ to 2.0 for this baseline. For these experiments, we use LLaMA-3-8B-Instruct as an example.
\begin{table}[]
    \small
    \centering
    \resizebox{\linewidth}{!}{
    \begin{tabular}{lcccc}
        \toprule
         & \multicolumn{4}{c}{$\pi_{ref}$=LLaMA-3-8B-Instruct}\\
         & \multicolumn{2}{c}{AlpacaEval} & MT & ArenaHard \\
         \cmidrule(lr){2-3} \cmidrule(lr){4-4} \cmidrule(lr){5-5} 
         & LR $\uparrow$ & WR $\uparrow$ & Score $\uparrow$ & WR $\uparrow$ \\
         \midrule
         Base       & 33.60 & 32.60 & 6.78 & 37.57 \\
         +DPO & 49.02 & 44.72 & 6.73 & \textbf{53.98} \\
         +\ourmethod & 60.77 & \textbf{58.29} & \textbf{7.19} & \underline{52.06} \\
         ~~w/ attn sink & \underline{61.14} & \underline{57.27} & \underline{6.93} & 51.64 \\
         ~~w/ len norm & \textbf{62.00} & 55.69 & 6.79 & 44.59 \\
         \bottomrule
    \end{tabular}}
    \caption{Ablation Results.}
    \label{tab:res_ablations}
    \vspace{-5pt}
\end{table}
Table~\ref{tab:res_ablations} shows that applying length normalization significantly hurts performance on MT-Bench and AreaHard. One speculated reason for this is the misalignment with the original RL objective. Instead of maximizing the total rewards of the entire rollout $y$, the dual-RL objective of objective~\ref{eq:twdpo_len_norm} (which is the same as objective~\ref{eq:attentionpo_rl} but with $|y|$ removed in both the reward function and the KL penalty) maximizes the average per-token reward of $y$, undermining the total rewards.

\section{Related Work}
\label{sec:related_work}

\paragraph{Preference Optimization.} 
Aligning LLMs with human preferences originally relied on RLHF via PPO \cite{rlhf2022, ppo2017}, which requires a separate reward model. Modern Preference Optimization (PO) methods bypass this by training the policy directly on preference pairs $(y_w, y_l)$ using contrastive objectives. For instance, RRHF \cite{rrhf2023} and SLiC-HF \cite{slichf2023} apply ranking losses, while DPO \cite{dpo2023} derives a reference-normalized objective from policy gradients. Subsequent variants introduce bounded objectives (IPO \cite{ipo2024}), pointwise formulations (KTO \cite{kto2024}), auxiliary SFT objectives (CPO \cite{cpo2024}, ORPO \cite{orpo2024}), length penalties (R-DPO \cite{rdpo2024}), or reference-free length normalization with reward margins (SimPO \cite{simpo2024}). However, these methods weight all response tokens equally, failing to capture token-level importance.

\paragraph{Token-level PO.} 
To enable fine-grained credit assignment, recent work explores token-level weighting. TDPO \cite{tdpo2024} incorporates token-level sequential KL terms. TIS-DPO \cite{tisdpo2025}, SePO \cite{sepo2025}, and cDPO \cite{cdpo2024} estimate token importance by contrasting probabilities between reference/negative models and oracle/positive models. TI-DPO \cite{tidpo2026} combines Gaussian priors with logit gradient norms. Alternatively, heuristic approaches like D$^2$PO \cite{earlier2025} decay weights based on token position. OTPO \cite{otpo2025} uses the euclidean distances between token hidden states to estimate token weights. Nonetheless, most of these methods either demand expensive auxiliary models \cite{tisdpo2025, sepo2025, cdpo2024} or rely on rigid, position-based heuristics \cite{earlier2025} rather than semantic content. Among the methods \cite{tdpo2024, tidpo2026, otpo2025} that share the same advantage as ours, which do not require an extra model for weight estimation and are adaptive to context, our method exhibits higher performance than theirs across models and benchmarks.

\paragraph{Attentions for credit assignment.} The idea of using attentions for fine-grained credit assignment has been explored in the context of reward shaping and found to work well. Prior work \cite{ferret2020, holmes2025} uses the reward model’s attentions for reward shaping, which densifies the rewards and boosts training efficiency in various RL environments. Attention-Based Credit \cite{chan2024} uses a similar approach to obtain dense rewards in the LLM RLHF (PPO \citep{schulman2017}) setting, which boosts efficiency and effectiveness. Our design choice of attention-based weights draws inspiration from these. The high-level motivation is that we use LLM as a pairwise judge (reward) model and use the attentions from the final verdict token (predicted reward) as the token weights.

\section{Conclusion}
We propose TwDPO---a token-weighted DPO objective, theoretically grounded on token-weighted RL---and \ourmethod---an instantiation of TwDPO that prompts the model to judge the response qualities and extract attentions from the model itself to weigh the response tokens. Experiment results show that \ourmethod~significantly improves the performance of the base model across AlpacaEval, MT-Bench, and ArenaHard, outperforming various prior PO methods. We hope \ourmethod~paves the way for highly-performant and efficient preference alignment with humans.

\section*{Limitations}

Due to computational constraints, we do not experiment with our method on larger models. In addition, our training and evaluation focus on the instruction-following task, but it can be applied to other more specialized domains such as reasoning or knowledge-based QA. Analysing the performance and generalizability of \ourmethod~on other models and tasks is an important future direction.

Moreover, the search space for the optimal choice of attention weights for TwDPO is large. For example, while we simply take the mean across attention heads, future work could investigate selecting a specific head or group of heads for weight extraction. However, we argue that such a choice of attention heads is likely model-dependent and may not transfer across settings.

\bibliography{custom}

@inproceedings{tidpo2026,
    author = {Ning Yang and Hai Lin and Yibo Liu and Baoliang Tian and Guoqing Liu and Haijun Zhang},
    title = {Token-Importance Guided Direct Preference Optimization},
    booktitle = {Proceedings of International Conference on Learning Representations},
    year = {2026}
}

@inproceedings{dpo2023,
    author = {Rafael Rafailov and Archit Sharma and Eric Mitchell and Stefano Ermon and Christopher D. Manning and Chelsea Finn},
    title = {Direct Preference Optimization: Your Language Model is Secretly a Reward Model},
    booktitle = {Advances in Neural Information Processing Systems},
    year = {2023}
}

@inproceedings{tisdpo2025,
    author = {Aiwei Liu and Haoping Bai and Zhiyun Lu and Yanchao Sun and Xiang Kong and Simon Wang and Jiulong Shan and Albin Madappally Jose and Xiaojiang Liu and Lijie Wen and Philip S. Yu and Meng Cao},
    title = {TIS-DPO: Token-level Importance Sampling for Direct Preference Optimization With Estimated Weights},
    booktitle = {Proceedings of International Conference on Learning Representations},
    year = {2025}
}

@inproceedings{sepo2025,
    author = {Kailai Yang and Zhiwei Liu and Qianqian Xie and Jimin Huang and Erxue Min and Sophia Ananiadou},
    title = {Selective Preference Optimization via Token-Level Reward Function Estimation},
    booktitle = {Proceedings of Empirical Methods in Natural Language Processing},
    year = {2025}
}

@inproceedings{tdpo2024,
    author = {Yongcheng Zeng and Guoqing Liu and Weiyu Ma and Ning Yang and Haifeng Zhang and Jun Wang},
    title = {Token-level Direct Preference Optimization},
    booktitle = {Proceedings of International Conference on Machine Learning},
    year = {2024}
}

@inproceedings{earlier2025,
    author = {Ruichen Shao and Bei Li and Gangao Liu and Yang Chen and Xiang Zhou and Jingang Wang and Xunliang Cai and Peng Li},
    title = {Earlier Tokens Contribute More: Learning Direct Preference Optimization From Temporal Decay Perspective},
    booktitle = {Proceedings of International Conference on Learning Representations},
    year = {2025},
}

@inproceedings{kto2024,
    author = {Kawin Ethayarajh and Winnie Xu and Niklas Muennighoff and Dan Jurafsky and Douwe Kiela},
    title = {KTO: Model Alignment as Prospect Theoretic Optimization},
    booktitle = {Proceedings of International Conference on Machine Learning},
    year = {2024}
}

@inproceedings{ipo2024,
    author = {Mohammad Gheshlaghi Azar and Mark Rowland and Bilal Piot and Daniel Guo and Daniele Calandriello and Michal Valko and Rémi Munos},
    title = {A General Theoretical Paradigm to Understand Learning from Human Preferences},
    booktitle = {Proceedings of the 27th International Conference on Artificial Intelligence and Statistics},
    year = {2024}
}

@inproceedings{rdpo2024,
    author = {Ryan Park and Rafael Rafailov and Stefano Ermon and Chelsea Finn},
    title = {Disentangling Length from Quality in Direct Preference Optimization},
    booktitle = {Proceedings of Findings of the Association for Computational Linguistics},
    year = {2024}
}

@inproceedings{orpo2024,
    author = {Jiwoo Hong and Noah Lee and James Thorne},
    title = {ORPO: Monolithic Preference Optimization without Reference Model},
    booktitle = {Proceedings of the 2024 Conference on Empirical Methods in Natural Language Processing},
    year = {2024}
}

@inproceedings{cpo2024,
    author = {Haoran Xu and Amr Sharaf and Yunmo Chen and Weiting Tan and Lingfeng Shen and Benjamin Van Durme and Kenton Murray and Young Jin Kim},
    title = {Contrastive Preference Optimization: Pushing the Boundaries of LLM Performance in Machine Translation},
    booktitle = {Proceedings of International Conference on Machine Learning},
    year = {2024}
}

@inproceedings{rrhf2023,
    author = {Zheng Yuan and Hongyi Yuan and Chuanqi Tan and Wei Wang and Songfang Huang and Fei Huang},
    title = {RRHF: Rank Responses to Align Language Models with Human Feedback without tears},
    booktitle = {Advances in Neural Information Processing Systems},
    year = {2023}
}

@inproceedings{simpo2024,
    author = {Yu Meng and Mengzhou Xia and Danqi Chen},
    title = {SimPO: Simple Preference Optimization with a Reference-Free Reward},
    booktitle = {Advances in Neural Information Processing Systems},
    year = {2024}
}

@article{slichf2023,
    author = {Yao Zhao and Rishabh Joshi and Tianqi Liu and Misha Khalman and Mohammad Saleh and Peter J. Liu},
    title = {SLiC-HF: Sequence Likelihood Calibration with Human Feedback},
    journal = {arXiv preprint arXiv:2305.10425},
    year = {2023}
}

@article{ppo2017,
    author = {John Schulman and Filip Wolski and Prafulla Dhariwal and Alec Radford and Oleg Klimov},
    title = {Proximal Policy Optimization Algorithms},
    journal = {arXiv preprint arXiv:1707.06347},
    year = {2017}
}

@article{rlhf2022,
    author = {Long Ouyang and Jeff Wu and Xu Jiang and Diogo Almeida and Carroll L. Wainwright and Pamela Mishkin and Chong Zhang and Sandhini Agarwal and Katarina Slama and Alex Ray and John Schulman and Jacob Hilton and Fraser Kelton and Luke Miller and Maddie Simens and Amanda Askell and Peter Welinder and Paul Christiano and Jan Leike and Ryan Lowe},
    title = {Training language models to follow instructions with human feedback},
    journal = {arXiv preprint arXiv:2203.02155},
    year = {2022}
}

@inproceedings{crowdsourced2024,
  title={From Crowdsourced Data to High-Quality Benchmarks: Arena-Hard and BenchBuilder Pipeline},
  author={Li, Tianle and Chiang, Wei-Lin and Frick, Evan and Dunlap, Lisa and Wu, Tianhao and Zhu, Banghua and Gonzalez, Joseph E and Stoica, Ion},
  booktitle={Proceedings of International Conference on Machine Learning},
  year={2025}
}

@misc{arenahard2024,
    title = {From Live Data to High-Quality Benchmarks: The Arena-Hard Pipeline},
    url = {https://lmsys.org/blog/2024-04-19-arena-hard/},
    author = {Tianle Li and Wei-Lin Chiang and Evan Frick and Lisa Dunlap and Banghua Zhu and Joseph E. Gonzalez and Ion Stoica},
    month = {April},
    year = {2024}
}

@misc{alpaca_eval,
  author = {Xuechen Li and Tianyi Zhang and Yann Dubois and Rohan Taori and Ishaan Gulrajani and Carlos Guestrin and Percy Liang and Tatsunori B. Hashimoto },
  title = {AlpacaEval: An Automatic Evaluator of Instruction-following Models},
  year = {2023},
  month = {5},
  publisher = {GitHub},
  journal = {GitHub repository},
  howpublished = {\url{https://github.com/tatsu-lab/alpaca_eval}}
}

@inproceedings{dubois2024length,
  title={Length-Controlled AlpacaEval: A Simple Way to Debias Automatic Evaluators},
  author={Dubois, Yann and Galambosi, Bal{\'a}zs and Liang, Percy and Hashimoto, Tatsunori B},
  booktitle={Proceedings of Conference on Language Modeling},
  year={2024}
}

@inproceedings{zheng2023,
    author = {Lianmin Zheng and Wei-Lin Chiang and Ying Sheng and Siyuan Zhuang and Zhanghao Wu and Yonghao Zhuang and Zi Lin and Zhuohan Li and Dacheng Li and Eric P. Xing and Hao Zhang and Joseph E. Gonzalez and Ion Stoica},
    title = {Judging LLM-as-a-Judge with MT-Bench and Chatbot Arena},
    booktitle = {Advances in Neural Information Processing Systems, Datasets and Benchmarks Track},
    year = {2023}
}

@article{llama32024,
    author = {Aaron Grattafiori and Abhimanyu Dubey and Abhinav Jauhri and Abhinav Pandey and Abhishek Kadian and Ahmad Al-Dahle and Aiesha Letman and Akhil Mathur and Alan Schelten and Alex Vaughan et al.},
    title = {The Llama 3 Herd of Models},
    journal = {arXiv preprint arXiv:2407.21783},
    year = {2024}
}

@article{gpt42023,
    author = {Aaron Hurst and Adam Lerer and Adam P. Goucher and Adam Perelman and Aditya Ramesh and Aidan Clark and AJ Ostrow and Akila Welihinda and Alan Hayes and Alec Radford et al.},
    title = {GPT-4 Technical Report},
    journal = {arXiv preprint arXiv:2303.08774},
    year = {2023}
}

@article{gpt4o2024,
    author = {Josh Achiam and Steven Adler and Sandhini Agarwal and Lama Ahmad and Ilge Akkaya and Florencia Leoni Aleman and Diogo Almeida and Janko Altenschmidt and Sam Altman and Shyamal Anadkat et al.},
    title = {GPT-4o System Card},
    journal = {arXiv preprint arXiv:2410.21276},
    year = {2024}
}

@inproceedings{attention2017,
  author={Ashish Vaswani and Noam Shazeer and Niki Parmar and Jakob Uszkoreit and Llion Jones and Aidan N. Gomez and Lukasz Kaiser and Illia Polosukhin},
  title={Attention Is All You Need},
  booktitle={Advances in Neural Information Processing Systems},
  year={2017}
}

@inproceedings{clark2019,
  author={Kevin Clark and Urvashi Khandelwal and Omer Levy and Christopher D. Manning},
  title={What Does BERT Look At?},
  booktitle={Proceedings of the 2019 ACL Workshop BlackboxNLP: Analyzing and Interpreting Neural Networks for NLP},
  year={2019}
}

@inproceedings{jain2019,
  author={Jain and Wallace},
  title={Attention is Not Explanation},
  booktitle={Proceedings of the 2019 Conference on Empirical Methods in Natural Language Processing and the 9th International Joint Conference on Natural Language Processing (EMNLP-IJCNLP)},
  year={2019}
}

@inproceedings{serrano2019,
  author={Sofia Serrano and Noah A. Smith},
  title={Is Attention Interpretable?},
  booktitle={Proceedings of the 57th Annual Meeting of the Association for Computational Linguistics},
  year={2019}
}

@inproceedings{vig2019,
  author={Jesse Vig},
  title={A Multiscale Visualization of Attention in the Transformer Model},
  booktitle={Proceedings of the 57th Annual Meeting of the Association for Computational Linguistics: System Demonstrations},
  year={2019}
}

@inproceedings{mccoy2021,
  author={R. Thomas McCoy and Ellie Pavlick and Tal Linzen},
  title={Right for the Wrong Reasons: Diagnosing Syntactic Heuristics in Natural Language Inference},
  booktitle={Proceedings of the 57th Annual Meeting of the Association for Computational Linguistics},
  year={2019}
}

@inproceedings{wiedemann2019,
  author={Gregor Wiedemann and Steffen Remus and Avi Chawla and Chris Biemann},
  title={Does BERT Make Any Sense? Interpretable Word Sense Disambiguation with Contextualized Embeddings},
  booktitle={Conference on Natural Language Processing},
  year={2019},
}

@inproceedings{attentionrollout2020,
    author = {Samira Abnar and Willem Zuidema},
    title = {Quantifying Attention Flow in Transformers},
    booktitle = {Proceedings of the 58th Annual Meeting of the Association for Computational Linguistics},
    year = {2020}
}

@article{bradley1952,
    author = {R. A. Bradley and M. E. Terry},
    title = {Rank analysis of incomplete block designs: I. the method of paired comparisons},
    journal = {Biometrika},
    volume = {39},
    issue = {3/4},
    pages = {324--345},
    year = {1952},
    doi = {https://doi.org/10.2307/2334029}
}

@inproceedings{xiao2024,
    author = {Guangxuan Xiao and Yuandong Tian and Beidi Chen and Song Han and Mike Lewis},
    title = {Efficient Streaming Language Models with Attention Sinks},
    booktitle = {Proceedings of International Conference on Machine Learning},
    year = {2024},
}

@inproceedings{huang2025,
    author = {Chengyu Huang and Tanya Goyal},
    title = {DCRM: A Heuristic to Measure Response Pair Quality in Preference Optimization},
    booktitle = {Findings of the Association for Computational Linguistics: EMNLP 2025},
    year = {2025}
}

@inproceedings{ultrafeedback2024,
    author = {Ganqu Cui and Lifan Yuan and Ning Ding and Guanming Yao and Bingxiang He and Wei Zhu and Yuan Ni and Guotong Xie and Ruobing Xie and Yankai Lin and Zhiyuan Liu and Maosong Sun},
    title = {UltraFeedback: Boosting Language Models with Scaled AI Feedback},
    booktitle = {Proceedings of International Conference on Machine Learning},
    year = {2024}
}

@inproceedings{adamw2019,
    author = {Ilya Loshchilov and Frank Hutter},
    title = {Decoupled Weight Decay Regularization},
    booktitle = {Proceedings of International Conference on Learning Representations},
    year = {2019}
}

@article{tunstall2023,
    author = {Lewis Tunstall and Edward Beeching and Nathan Lambert and Nazneen Rajani and Kashif Rasul and Younes Belkada and Shengyi Huang and Leandro von Werra and Clémentine Fourrier and Nathan Habib, et al},
    title = {Zephyr: Direct distillation of lm alignment.},
    journal = {arXiv preprint arXiv:2310.16944},
    year = {2023}
}

@inproceedings{ultrachat2023,
    author = {Ning Ding and Yulin Chen and Bokai Xu and Yujia Qin and Shengding Hu and Zhiyuan Liu and Maosong Sun and Bowen Zhou},
    title = {Enhancing Chat Language Models by Scaling High-quality Instructional Conversations},
    booktitle = {Proceedings of the 2023 Conference on Empirical Methods in Natural Language Processing},
    year = {2023}
}

@article{mistral2023,
    author = {Albert Q. Jiang and Alexandre Sablayrolles and Arthur Mensch and Chris Bamford and Devendra Singh Chaplot and Diego de las Casas and Florian Bressand and Gianna Lengyel and Guillaume Lample and Lucile Saulnier and Lélio Renard Lavaud and Marie-Anne Lachaux and Pierre Stock and Teven Le Scao and Thibaut Lavril and Thomas Wang and Timothée Lacroix and William El Sayed},
    title = {Mistral 7B},
    journal = {arXiv preprint arXiv:2310.06825},
    year = {2023}
}

@inproceedings{llmblender2023,
    author = {Dongfu Jiang and Xiang Ren and Bill Yuchen Lin},
    title = {LLM-Blender: Ensembling Large Language Models with Pairwise Ranking and Generative Fusion},
    booktitle = {Proceedings of the 61st Annual Meeting of the Association for Computational Linguistics (Volume 1: Long Papers)},
    year = {2023}
}

@article{cdpo2024,
    author = {Zicheng Lin and Tian Liang and Jiahao Xu and Qiuzhi Lin and Xing Wang and Ruilin Luo and Chufan Shi and Siheng Li and Yujiu Yang and Zhaopeng Tu},
    title = {Critical Tokens Matter: Token-Level Contrastive Estimation Enhances LLM's Reasoning Capability},
    journal = {arXiv preprint arXiv:2411.19943},
    year = {2024}
}

@inproceedings{otpo2025,
    author = {Meng Li and Guangda Huzhang and Haibo Zhang and Xiting Wang and Anxiang Zeng},
    title = {Optimal Transport-Based Token Weighting scheme for Enhanced Preference Optimization},
    booktitle = {Proceedings of the 63th Annual Meeting of the Association for Computational Linguistics},
    year = {2025},
}

@inproceedings{chan2024,
    author = {Alex J. Chan and Hao Sun and Samuel Holt and Mihaela van der Schaar},
    title = {Dense Reward for Free in Reinforcement Learning from Human Feedback},
    booktitle = {Proceedings of International Conference on Machine Learning},
    year = {2024}
}

@inproceedings{ferret2020,
    author = {Johan Ferret and Raphaël Marinier and Matthieu Geist and Olivier Pietquin},
    title = {Self-Attentional Credit Assignment for Transfer in Reinforcement Learning},
    booktitle = {Proceedings of International Joint Conferences on Artificial Intelligence},
    year = {2020},
}

@article{holmes2025,
    author = {Ian Holmes and Min Chi},
    title = {Attention-Based Reward Shaping for Sparse and Delayed Rewards},
    journal = {arXiv preprint arXiv:2505.10802},
    year = {2025}
}

@article{schulman2017,
    author = {John Schulman and Filip Wolski and Prafulla Dhariwal and Alec Radford and Oleg Klimov},
    title = {Proximal Policy Optimization Algorithms},
    journal = {arXiv preprint arXiv:1707.06347},
    year = {2017}
}

\clearpage
\appendix

\section{Extended Related Work}

\paragraph{Preference Optimization.} The alignment with human preferences and values is critical for LLM-generated output. Early work \cite{rlhf2022} trains LLM using policy-gradient RL-algorithm such as PPO \cite{ppo2017}. However, these methods require training a separate reward model on annotated preference pairs to provide RL rewards. Preference Optimization (PO) methods avoid the need for an extra reward model and train the main policy directly on the preference pairs to learn the reward landscape. On a high level, PO methods use a contrastive learning objective between a preferred response $y_w$ and a dispreferred response $y_l$. For example, RRHF~\cite{rrhf2023} and SLiC-HF~\cite{slichf2023} use ranking losses. DPO~\cite{dpo2023} drives its training objective from policy gradient, which normalizes the model's probability with the reference model's probability. IPO \cite{ipo2024} uses a bounded objective in contrast to DPO. KTO \cite{kto2024} avoids the pairwise assumption and uses a pointwise formulation. CPO \cite{cpo2024} and ORPO \cite{orpo2024} add an SFT term in addition to the contrastive term. R-DPO \cite{rdpo2024} adds an extra term to the original DPO objective to avoid the length exploitation. SimPO \cite{simpo2024} modifies DPO by removing dependencies on the reference model, incorporating length normalization, and adding a reward margin term. However, all the methods above assign equal weights to every response token, which cannot account for the fine-grained importance of each token to the overall response quality.

\paragraph{Token-level PO.} Various attempts are made to enable fine-grained, token-level credit assignment during PO training. TDPO \cite{tdpo2024} is grounded on token-level policy gradient methods. It adds two sequence KL terms to the original DPO objective. TIS-DPO \cite{tisdpo2025} and cDPO \cite{cdpo2024} estimate the token importance via contrasting the probability to produce the token from a positive policy with the one from a negative policy, given the same prefix. The positive and negative policies are induced either from prompts or training. SePO \cite{sepo2025} takes a similar approach, but it trains an oracle model with DPO on a subset of the original training set and contrasts the probability from the oracle against the one from the reference model. TI-DPO \cite{tidpo2026} estimates the token weights by combining a pre-defined Gaussian prior distribution with the gradient norm of the last token's maximum logit with respect to each token. Certain methods use heuristics to assign weights. For example, D$^2$PO \cite{earlier2025} gives earlier tokens more weight than later ones. However, many of these methods require training extra models to estimate weights.
% In addition, the use of gradients or probability ratios for credit assignments undermines interpretability.
The rest compute weights using heuristic functions, which are purely based on token positions rather than the tokens themselves.

\paragraph{Attention Weights.} Early studies on transformers \cite{attention2017} indicate that specific attention heads capture linguistic properties like coreference resolution, framing attention weights as transparent windows into model reasoning \cite{clark2019}. However, subsequent research challenges this notion by demonstrating that attention distributions can be manipulated without altering model predictions, implying that attention correlates with but does not cause model outputs \cite{jain2019, serrano2019}. Counter-counter-arguments suggest attention remains a useful, albeit incomplete, diagnostic signal when contextualized \cite{wiedemann2019}.  While individual attention heads specialize \cite{vig2019}, they operate redundantly. Furthermore, attention patterns are frequently associated with superficial heuristics rather than deep semantic understanding \cite{mccoy2021}. When aggregated across multiple layers with residual connections, the aggregated attentions show a higher correlation with gold token importance scores, compared with raw attentions \cite{attentionrollout2020}. These findings suggest that attention could potentially provide a complementary view of information flow. In summary, whether attention is interpretable is still under debate. As such, in this work, we do not explicitly assume the interpretability of attentions.

\section{Theoretical Analysis and Error Bounds}
\label{sec:error_bound}

Because the sequence-level expectation $\mathbb{E}_{y \sim \pi}$ in the TwDPO objective is coupled with token-weighted log-probabilities, deriving an exact analytical closed-form solution for the optimal policy is intractable without invoking complex token-level dynamic programming. In this section, we establish a formal theoretical guarantee for TwDPO by framing it as a bounded perturbation around the standard Direct Preference Optimization (DPO) objective \citep{dpo2023}. We first bound the divergence between the true, intractable TwDPO optimal policy $\pi_{\text{opt}}$ and the standard DPO optimal policy $\pi_{\text{DPO}}$. We then bound the distance between our empirical autoregressive heuristic policy $\tilde{\pi}^*$ and $\pi_{\text{DPO}}$, concluding with a unified Total Variation (TV) distance bound between the empirical heuristic and the true optimum.

\subsection{Objective Formulation and Perturbation Setup}

Let $\mathcal{D}$ denote the dataset of prompts $x$, and let $y = (y^1, y^2, \dots, y^{|y|})$ represent a generated sequence of length $|y|$. For a given prompt $x$, policy $\pi$, and reference policy $\pi_{\text{ref}}$, the TwDPO objective $J_{\text{TwDPO}}(\pi)$ modifies the standard sequence-level KL penalty by weighting individual token steps using attention weights $a^t$:
\begin{equation}
\small
\begin{split}
&J_{\text{TwDPO}}(\pi) = \mathbb{E}_{x \sim \mathcal{D}} \mathbb{E}_{y \sim \pi(\cdot|x)}\\&~~~~~~~~~~ \left[ r(x,y) - \beta |y| \sum_{t=1}^{|y|} a^t \log \frac{\pi(y^t \mid x, y^{<t})}{\pi_{\text{ref}}(y^t \mid x, y^{<t})} \right],
\end{split}
\end{equation}
where $\beta > 0$ is the global scaling parameter. By definition, the token attention weights are normalized such that $\sum_{t=1}^{|y|} a^t = 1$. Let us define the effective token weights as $w^t = |y|a^t$, satisfying $\frac{1}{|y|}\sum_{t=1}^{|y|} w^t = 1$.

We parameterize the deviation of these attention weights from a uniform distribution as a perturbation vector $\epsilon^t$:
\begin{equation}
w^t = 1 + \epsilon^t, \quad \text{where} \quad \sum_{t=1}^{|y|} \epsilon^t = 0.
\end{equation}

\begin{assumption}[Bounded Attention Deviation]
\label{assump:bounded_weights}
There exists a constant $\delta \ge 0$ such that the maximum token-level attention perturbation is strictly bounded for all sequences:
\begin{equation}
\max_{t} |\epsilon^t| \le \delta.
\end{equation}
\end{assumption}

\begin{assumption}[Bounded Trust Region Log-Ratios]
\label{assump:bounded_log_ratios}
The log-probability ratio between any valid policy $\pi$ under consideration and the reference policy $\pi_{\text{ref}}$ is bounded at the token level by a finite constant $C > 0$:
\begin{equation}
\max_{t} \left| \log \frac{\pi(y^t \mid x, y^{<t})}{\pi_{\text{ref}}(y^t \mid x, y^{<t})} \right| \le C.
\end{equation}
\end{assumption}

Using this formulation, we decompose the TwDPO objective into the standard DPO objective $J_{\text{DPO}}(\pi)$ and a descriptive perturbation term:
\begin{equation}
\small
\begin{split}
J_{\text{TwDPO}}(\pi) &= \mathbb{E}_{x \sim \mathcal{D}} \mathbb{E}_{y \sim \pi(\cdot|x)}\\& \left[ r(x,y) - \beta \sum_{t=1}^{|y|} (1 + \epsilon^t) \log \frac{\pi(y^t \mid x, y^{<t})}{\pi_{\text{ref}}(y^t \mid x, y^{<t})} \right] \nonumber \\
&= J_{\text{DPO}}(\pi) - \beta \mathbb{E}_{x \sim \mathcal{D}} \mathbb{E}_{y \sim \pi(\cdot|x)} \left[ R_{\epsilon}(\pi; x, y) \right],
\end{split}
\end{equation}
where the point-wise sequence perturbation $R_{\epsilon}(\pi; x, y)$ is defined as:
\begin{equation}
R_{\epsilon}(\pi; x, y) = \sum_{t=1}^{|y|} \epsilon^t \log \frac{\pi(y^t \mid x, y^{<t})}{\pi_{\text{ref}}(y^t \mid x, y^{<t})}.
\end{equation}

\subsection{Bounding the Objective Discrepancy}

We now bound the magnitude of the expected perturbation term. Under Assumptions \ref{assump:bounded_weights} and \ref{assump:bounded_log_ratios}, the sequence-level perturbation can be upper-bounded via the triangle inequality:
\begin{equation}
\begin{split}
    |R_{\epsilon}(\pi; x, y)| &\le \sum_{t=1}^{|y|} |\epsilon^t| \cdot \left| \log \frac{\pi(y^t \mid x, y^{<t})}{\pi_{\text{ref}}(y^t \mid x, y^{<t})} \right| \\&\le \sum_{t=1}^{|y|} \delta C = |y| \delta C.
\end{split}
\end{equation}

Taking the expectation over the prompt and sequence distributions, the discrepancy between the two objectives is strictly bounded by:
\begin{equation}
\small
\left| J_{\text{TwDPO}}(\pi) - J_{\text{DPO}}(\pi) \right| \le \beta \delta C \mathbb{E}_{x \sim \mathcal{D}} \mathbb{E}_{y \sim \pi(\cdot|x)} [|y|].
\end{equation}
For ease of notation, let $B(\pi) = \beta \delta C \mathbb{E}_{\pi}[|y|]$ denote the policy-dependent bound constraint.

\subsection{Distance Bound for True Optima}

Let $\pi_{\text{opt}} = \arg\max_{\pi} J_{\text{TwDPO}}(\pi)$ be the true, intractable optimal policy under the token-weighted framework, and let $\pi_{\text{DPO}} = \arg\max_{\pi} J_{\text{DPO}}(\pi)$ be the analytical optimal policy for the standard DPO objective. 

From the exact mathematical properties of KL-constrained reinforcement learning objectives, the suboptimality gap of any arbitrary policy $\pi$ evaluated under $J_{\text{DPO}}$ is exactly equal to its generalized KL divergence from the optimal DPO policy:
\begin{equation}
\label{eq:dpo_identity}
\small
\begin{split}
    &J_{\text{DPO}}(\pi_{\text{DPO}}) - J_{\text{DPO}}(\pi) \\&= \beta \mathbb{E}_{x \sim \mathcal{D}} \left[ D_{\text{KL}}(\pi(\cdot \mid x) \parallel \pi_{\text{DPO}}(\cdot \mid x)) \right].
\end{split}
\end{equation}

Evaluating the true optimal policy $\pi_{\text{opt}}$ using the identity in Equation \ref{eq:dpo_identity} yields:
\begin{equation}
\label{eq:kl_target}
\small
\begin{split}
&J_{\text{DPO}}(\pi_{\text{DPO}}) - J_{\text{DPO}}(\pi_{\text{opt}}) \\&= \beta \mathbb{E}_{x \sim \mathcal{D}} \left[ D_{\text{KL}}(\pi_{\text{opt}}(\cdot \mid x) \parallel \pi_{\text{DPO}}(\cdot \mid x)) \right].
\end{split}
\end{equation}

By the definition of functional optimality, $\pi_{\text{opt}}$ maximizes $J_{\text{TwDPO}}$, meaning $J_{\text{TwDPO}}(\pi_{\text{opt}}) \ge J_{\text{TwDPO}}(\pi_{\text{DPO}})$. We construct a double inequality chain by introducing the objective boundaries $B(\pi_{\text{opt}})$ and $B(\pi_{\text{DPO}})$:
\begin{equation}
\small
\begin{split}
    J_{\text{DPO}}(\pi_{\text{opt}}) + B(\pi_{\text{opt}}) &\ge J_{\text{TwDPO}}(\pi_{\text{opt}}) \\&\ge J_{\text{TwDPO}}(\pi_{\text{DPO}}) \\&\ge J_{\text{DPO}}(\pi_{\text{DPO}}) - B(\pi_{\text{DPO}}).
\end{split}
\end{equation}

Rearranging the outermost components of this inequality gives an upper bound on the DPO objective gap:
\begin{equation}
\small
J_{\text{DPO}}(\pi_{\text{DPO}}) - J_{\text{DPO}}(\pi_{\text{opt}}) \le B(\pi_{\text{DPO}}) + B(\pi_{\text{opt}}).
\end{equation}

Substituting this directly back into the identity in Equation \ref{eq:kl_target}, we obtain:
\begin{equation}
\small
\begin{split}
    \beta \mathbb{E}_{x \sim \mathcal{D}} \left[ D_{\text{KL}}(\pi_{\text{opt}}(\cdot \mid x) \parallel \pi_{\text{DPO}}(\cdot \mid x)) \right] \\ \le \beta \delta C \left( \mathbb{E}_{\pi_{\text{DPO}}}[|y|] + \mathbb{E}_{\pi_{\text{opt}}}[|y|] \right).
\end{split}
\end{equation}

Dividing both sides by the regularization parameter $\beta$ establishes our first localized distance guarantee.

\begin{lemma}[Optima Distance Bound]
\label{lemma:optima_bound}
The expected KL divergence between the true optimal token-weighted policy $\pi_{\text{opt}}$ and the standard DPO policy $\pi_{\text{DPO}}$ is linearly bounded by the maximum attention perturbation $\delta$:
\begin{equation}
\label{eq:opt_dpo_bound}
\small
\begin{split}
    \mathbb{E}_{x \sim \mathcal{D}} \left[ D_{\text{KL}}(\pi_{\text{opt}}(\cdot \mid x) \parallel \pi_{\text{DPO}}(\cdot \mid x)) \right] \\ \le \delta C \left( \mathbb{E}_{\pi_{\text{DPO}}}[|y|] + \mathbb{E}_{\pi_{\text{opt}}}[|y|] \right).
\end{split}
\end{equation}
\end{lemma}

\subsection{Bounding the Autoregressive Heuristic Policy}

We now address our empirical heuristic policy, denoted as $\tilde{\pi}^*$. While $\tilde{\pi}^*$ is derived via a localized parameterization mapping, it is generated sequentially in an autoregressive fashion, ensuring that $\tilde{\pi}^*(y \mid x) = \prod_{t=1}^{|y|} \tilde{\pi}^*(y^t \mid x, y^{<t})$ constitutes a mathematically proper, normalized probability distribution over sequence space ($\sum_{y} \tilde{\pi}^*(y \mid x) = 1$). 

By definition, our empirical token-weighted optimal parameterization satisfies the equality:
\begin{equation}
\small
\sum_{t=1}^{|y|} w^t \log \frac{\tilde{\pi}^*(y^t \mid x, y^{<t})}{\pi_{\text{ref}}(y^t \mid x, y^{<t})} = \frac{1}{\beta} r(x,y) - \log Z(x).
\end{equation}
Expanding $w^t = 1 + \epsilon^t$ yields:
\begin{equation}
\label{eq:heuristic_expanded}
\small
\log \frac{\tilde{\pi}^*(y \mid x)}{\pi_{\text{ref}}(y \mid x)} + R_{\epsilon}(\tilde{\pi}^*; x, y) = \frac{1}{\beta} r(x,y) - \log Z(x).
\end{equation}

Recall that the standard sequence-level DPO optimal policy satisfies:
\begin{equation}
\label{eq:dpo_canonical}
\small
\log \frac{\pi_{\text{DPO}}(y \mid x)}{\pi_{\text{ref}}(y \mid x)} = \frac{1}{\beta} r(x,y) - \log Z_{\text{DPO}}(x).
\end{equation}
Subtracting Equation \ref{eq:dpo_canonical} from Equation \ref{eq:heuristic_expanded} isolates the log-ratio between our heuristic policy and the standard DPO policy:
\begin{equation}
\label{eq:log_ratio_diff}
\small
\log \frac{\tilde{\pi}^*(y \mid x)}{\pi_{\text{DPO}}(y \mid x)} = \log Z_{\text{DPO}}(x) - \log Z(x) - R_{\epsilon}(\tilde{\pi}^*; x, y).
\end{equation}

Let $\Delta Z(x) = \log Z_{\text{DPO}}(x) - \log Z(x)$. Taking the expectation of Equation \ref{eq:log_ratio_diff} under $\tilde{\pi}^*$ yields the forward KL divergence:
\begin{equation}
\label{eq:forward_kl}
\small
D_{\text{KL}}(\tilde{\pi}^* \parallel \pi_{\text{DPO}}) = \Delta Z(x) - \mathbb{E}_{y \sim \tilde{\pi}^*}\left[ R_{\epsilon}(\tilde{\pi}^*; x, y) \right].
\end{equation}
Similarly, taking the expectation of Equation \ref{eq:log_ratio_diff} under the standard DPO policy yields the reverse KL divergence:
\begin{equation}
\label{eq:reverse_kl}
\small
-D_{\text{KL}}(\pi_{\text{DPO}} \parallel \tilde{\pi}^*) = \Delta Z(x) - \mathbb{E}_{y \sim \pi_{\text{DPO}}}\left[ R_{\epsilon}(\tilde{\pi}^*; x, y) \right].
\end{equation}

Subtracting Equation \ref{eq:reverse_kl} from Equation \ref{eq:forward_kl} cancels the intractable partition scaling term $\Delta Z(x)$:
\begin{equation}
\label{eq:kl_sum}
\small
\begin{split}
    &D_{\text{KL}}(\tilde{\pi}^* \parallel \pi_{\text{DPO}}) + D_{\text{KL}}(\pi_{\text{DPO}} \parallel \tilde{\pi}^*) \\&= \mathbb{E}_{y \sim \pi_{\text{DPO}}}\left[ R_{\epsilon}(\tilde{\pi}^*; x, y) \right] - \mathbb{E}_{y \sim \tilde{\pi}^*}\left[ R_{\epsilon}(\tilde{\pi}^*; x, y) \right].
\end{split}
\end{equation}

Since the reverse KL divergence is strictly non-negative ($D_{\text{KL}}(\pi_{\text{DPO}} \parallel \tilde{\pi}^*) \ge 0$), we drop it to establish an upper bound, and apply the triangle inequality across expectations:
\begin{equation}
\small
\begin{split}
    D_{\text{KL}}(\tilde{\pi}^* \parallel \pi_{\text{DPO}}) &\le \left| \mathbb{E}_{y \sim \pi_{\text{DPO}}}\left[ R_{\epsilon}(\tilde{\pi}^*; x, y) \right] \right| + \\&~~~~~~~~~~~~ \left| \mathbb{E}_{y \sim \tilde{\pi}^*}\left[ R_{\epsilon}(\tilde{\pi}^*; x, y) \right] \right| \nonumber \\
&\le \delta C \mathbb{E}_{y \sim \pi_{\text{DPO}}}[|y|] + \delta C \mathbb{E}_{y \sim \tilde{\pi}^*}[|y|].
\end{split}
\end{equation}

Taking the expectation across the entire prompt distribution $\mathcal{D}$ establishes our second localization lemma.

\begin{lemma}[Heuristic Distance Bound]
\label{lemma:heuristic_bound}
The expected KL divergence between the empirical autoregressive heuristic policy $\tilde{\pi}^*$ and the closed-form standard DPO policy $\pi_{\text{DPO}}$ is strictly bounded by:
\begin{equation}
\label{eq:nop_dpo_bound}
\small
\begin{split}
    \mathbb{E}_{x \sim \mathcal{D}} \left[ D_{\text{KL}}(\tilde{\pi}^*(\cdot \mid x) \parallel \pi_{\text{DPO}}(\cdot \mid x)) \right] \\\le \delta C \left( \mathbb{E}_{\pi_{\text{DPO}}}[|y|] + \mathbb{E}_{\tilde{\pi}^*}[|y|] \right).
\end{split}
\end{equation}
\end{lemma}

\subsection{Unified Suboptimality Guarantee via Total Variation Distance}

We now unify Lemma \ref{lemma:optima_bound} and Lemma \ref{lemma:heuristic_bound} to bound the final distance between our empirical heuristic policy $\tilde{\pi}^*$ and the true intractable optimal policy $\pi_{\text{opt}}$. We map our KL bounds to Total Variation (TV) distance using Pinsker's inequality, which states that for any distributions $P$ and $Q$, $D_{\text{TV}}(P, Q) \le \sqrt{\frac{1}{2} D_{\text{KL}}(P \parallel Q)}$.

Applying Pinsker's inequality to Lemma \ref{lemma:optima_bound} and Lemma \ref{lemma:heuristic_bound} gives:
\begin{equation}
\label{eq:tv_opt}
\small
\begin{split}
&\mathbb{E}_{x \sim \mathcal{D}} [D_{\text{TV}}(\pi_{\text{opt}}, \pi_{\text{DPO}})]\\
&\le \mathbb{E}_{x \sim \mathcal{D}}\biggl[\sqrt{\frac{1}{2}D_{\text{KL}}\left(\pi_{\text{opt}}(\cdot \mid x) \parallel \pi_{\text{DPO}}(\cdot \mid x)  \right)}\biggr]\\
&\le \mathbb{E}_{x \sim \mathcal{D}}\biggl[\sqrt{\frac{\delta C}{2} \left( \mathbb{E}_{y\sim\pi_{\text{DPO}}}[|y|] + \mathbb{E}_{y\sim\pi_{\text{opt}}}[|y|] \right)}\biggr]\\
&\le \sqrt{\frac{\delta C}{2} \left( \mathbb{E}_{\pi_{\text{DPO}}}[|y|] + \mathbb{E}_{\pi_{\text{opt}}}[|y|] \right)},
\end{split}
\end{equation}
where the first inequality comes from Pinsker's inequality, the second comes from equation~\ref{eq:opt_dpo_bound} with $\mathcal{D}$ containing only a single query $x$, the third comes from Jensen's inequality.

Similarly, applying Pinsker's inequality, equation~\ref{eq:nop_dpo_bound} with $\mathcal{D}$ containing only a single query $x$, and Jensen's inequality, we get
\begin{equation}
\small
\label{eq:tv_heur}
\begin{split}
&\mathbb{E}_{x \sim \mathcal{D}} [D_{\text{TV}}(\tilde{\pi}^*, \pi_{\text{DPO}})] \le \sqrt{\frac{\delta C}{2} \left( \mathbb{E}_{\pi_{\text{DPO}}}[|y|] + \mathbb{E}_{\tilde{\pi}^*}[|y|] \right)}.
\end{split}
\end{equation}

Finally, by utilizing the triangle inequality property of Total Variation distances, we bound the direct distance between the empirical heuristic and the true functional optimum:
\begin{equation}
\small
D_{\text{TV}}(\pi_{\text{opt}}, \tilde{\pi}^*) \le D_{\text{TV}}(\pi_{\text{opt}}, \pi_{\text{DPO}}) + D_{\text{TV}}(\tilde{\pi}^*, \pi_{\text{DPO}}).
\end{equation}

Taking expectations over $\mathcal{D}$ and substituting Equations \ref{eq:tv_opt} and \ref{eq:tv_heur} leads directly to our main result.

\begin{theorem}[TwDPO Suboptimality Consistancy]
Under Assumptions \ref{assump:bounded_weights} and \ref{assump:bounded_log_ratios}, the expected Total Variation distance between the empirical autoregressive heuristic policy $\tilde{\pi}^*$ and the true, intractable token-weighted optimal policy $\pi_{\text{opt}}$ is strictly bounded by $\mathcal{O}(\sqrt{\delta})$:
\begin{equation}
\small
\begin{split}
    &\mathbb{E}_{x \sim \mathcal{D}} [D_{\text{TV}}(\pi_{\text{opt}}(\cdot \mid x), \tilde{\pi}^*(\cdot \mid x))] \le \sqrt{\frac{\delta C}{2}}\\&~~~~ \left( \sqrt{\mathbb{E}_{\pi_{\text{DPO}}}[|y|] + \mathbb{E}_{\pi_{\text{opt}}}[|y|]} + \sqrt{\mathbb{E}_{\pi_{\text{DPO}}}[|y|] + \mathbb{E}_{\tilde{\pi}^*}[|y|]} \right).
\end{split}
\end{equation}
\end{theorem}

This theorem provides a rigorous theoretical safety net for TwDPO. It demonstrates that while a clean closed-form sequence-level derivation is structurally impossible due to token-weight expectation coupling, the empirical objective remains mathematically consistent. As long as the attention variance deviation from uniform is controlled ($\delta \to 0$), the empirical policy behaves predictably and stays bounded within a localized neighborhood of the true mathematical optimum.

\section{Gradient Analysis}
\label{app:gradient_analysis}
The gradient of objective~\ref{eq:twdpo} is:

\begin{equation}
\small
\begin{split}
    \nabla_\theta\mathcal{L}&=-\beta\mathbb{E}_{(x,y_w,y_l)\sim \mathcal{D}}\biggl[\sigma \biggl(r'_\theta(x,y_l)-r'_\theta(x,y_w) \biggr) \\
    &\biggl(|y_w|\sum_{t=1}^{|y_w|}a_{w}^{t}\nabla_{\theta}log\pi_{\theta}(y_w^t|x,y_w^{<t})-\\
    &~~~|y_l|\sum_{t=1}^{|y_l|}a_{l}^{t}\nabla_{\theta}log\pi_{\theta}(y_l^t|x,y_l^{<t})\biggr) \biggr],
\end{split}
\end{equation}
where $r'_\theta(x,y)=\beta |y|\sum_{t=1}^{|y|} a^{t} \log \frac{\pi_{\theta}(y^t|x,y^{<t})}{\pi_{ref}(y^t|x,y^{<t})}$.

Similar to DPO, the $\sigma(r'_\theta(x,y_l)-r'_\theta(x,y_w))$ term scales up our combined gradient when the reward margin between $y_w$ and $y_l$ is smaller or negative. The main difference from DPO is that we weigh the gradient of each token at position $t$ by $a^t$, so that higher-weighted tokens get more gradient updates and the model learns more on these tokens.

\section{Note on the weighted log probability term}
\label{app:weighted_prob_term}
Despite the use of notation $\tilde{\pi}$, our original derivation does not require the weighted probability terms $\tilde{\pi}$ and $\tilde{\pi}_{ref}$ to be proper distributions. Nonetheless, we can normalize the weighted likelihood to form a proper distribution, and the near-optimal policy, as well as its error bound, remains the same. The true optimal policy remains intractable because, in contrast to the original RL objective, the token-weighted RL objective cannot be expressed as a proper KL divergence between the sampling policy and the optimal policy.

In the following, we show how to make the weighted log-likelihood a proper distribution.

Define $\log \bar{\pi}(y|x) = \log \tilde{\pi}(y|x) - \log C^1(x)$. Then $\bar{\pi}(y|x) = \tilde{\pi}(y|x) / C^1(x)$. Setting $C^1(x) = \sum_{y}  \tilde{\pi}(y|x)$ makes $\bar{\pi}(y|x)$ a proper distribution. We can define $\bar{\pi}_{ref}(y|x)$ in a similar way. Setting $C^2(x) = \sum_{y}  \tilde{\pi}_{ref}(y|x)$ makes it a proper distribution.

Now denote $\tilde{Q} = \log \tilde{\pi}(y|x) / \tilde{\pi}_{ref}(y|x)$ and $\bar{Q} = \log \bar{\pi}(y|x) / \bar{\pi}_{ref}(y|x)$. Then the TwDPO objective in equation~\ref{eq:attentionpo_rl} is equivalent to

\begin{equation*}
\small
\begin{split}
& \max_{\pi}\mathbb{E}_{x\sim\mathcal{D},y\sim\pi(\cdot|x)}\biggl[r(x,y) -\beta \tilde{Q} \biggr]\\
& = \max_{\pi}\mathbb{E}_{x\sim\mathcal{D},y\sim\pi(\cdot|x)}\biggl[r(x,y) -\beta \tilde{Q} - \\
&~~~~~~~~~~~~~~~~~~~~~~~~~~~~~~~~~~\log C^1(x) - \log C^2(x) \biggr]\\
& = \max_{\pi}\mathbb{E}_{x\sim\mathcal{D},y\sim\pi(\cdot|x)}\biggl[r(x,y) -\beta \bar{Q} \biggr]\\
& = \max_{\pi}\mathbb{E}_{x\sim\mathcal{D},y\sim\pi(\cdot|x)} \biggl[ \frac{\bar{\pi}(y|x)}{\bar{\pi}^*(y|x)}  - \log \bar{Z}(x) \biggr] \quad(1)^* 
\end{split}
\end{equation*}
where
\begin{equation*}
\small
\begin{split}
    \bar{\pi}^*(y|x) = \frac{\bar{\pi}_{ref}(y|x) exp(\frac{1}{\beta}r(x, y))}{\bar{Z}(x)} \quad \quad (2)^*
\end{split}
\end{equation*}
and
\begin{equation*}
\small
\begin{split}
        \bar{Z}(x) = \sum_{y} \bar{\pi}_{ref}(y|x) exp(\frac{1}{\beta}r(x, y))
\end{split}
\end{equation*}

Note that, for every y, $\bar{\pi}_{ref}(y|x)$ is just the original $\tilde{\pi}_{ref}(y|x)$ multiplied by $C^2(x)$. Expanding it makes the $C^2(x)$ in the numerator and denominator in $(2)^*$ cancel out.

Therefore, the new near-optimum policy $\bar{\pi}^*(y|x)$ is equivalent to the original near-optimum policy in Equation~\ref{eq:weighted_pi} and has the same error bound. The new reward is the same as in Equation~\ref{eq:attention_po_reward}, but there is an extra $-\log C^1(x) - \log C^2(x)$ term, which cancels out in pairwise subtraction, and leads to the same optimization objective~\ref{eq:reward_prob}.

So we can make the weighted likelihood a proper distribution and it leads to the exact same final objective, but we argue that the true optimal solution is still intractable because the expectation in $(1)^*$ is over the original policy $y \sim \pi(\cdot|x)$, not the new distribution $y\sim \bar{\pi}(\cdot|x)$, and so $(1)^*$ is not a proper KL divergence.

\section{Token Matching}
\label{app:token_matching}

We note that certain attention scores may not be matched to any tokens, and certain tokens may not be assigned an attention weight. This is because when we extract the attention weights in \S~\ref{sec:method_awe}, the response $y$ is tokenized with the pairwise judge prompt as the surrounding context. During training, $y$ is tokenized with a different surrounding context, and consequently, the first and last several tokens do not match the original tokens extracted from the pairwise judge prompt. In case no perfect match is found, we use the "edit\_distance" Python library to match the attention weight to the response tokens during training. For tokens without a match, we assign a weight of 0. We observe that in most cases, more than 95\% of the tokens are matched to attention weights.

\section{Hyperparameters}
\label{app:hyperparameters}

\begin{table}[htp!]
    \centering
    \begin{tabular}{lc}
        \toprule
        Parameter & Value \\
        \midrule
         Max Context Length & 2048 \\
         Max Prompt Length & 1800 \\
         Learning Rate & 1e-6 \\
         $\beta$ & 5e-3 \\
         Batch Size & 32 \\
         \# Epochs & 1 \\
         \# Train Examples & $\sim$60K \\
         \# Validation Examples & $\sim$2K \\
         Scheduler & Cosine \\
         Warmup Ratio & 0.1 \\
         Optimizer & AdamW \\
         \bottomrule
    \end{tabular}
    \caption{Training Hyperparameters}
    \label{tab:training_hyperparameters}
\end{table}

\paragraph{Training.} We tune our training $\beta$ in the range $\{\text{1e-3, 5e-3, 1e-2, 2e-2}\}$ and show the final training hyperparameters for \ourmethod for both LLaMA-3-8B-Base-SFT and LLaMA-3-8B-Instruct in Figure~\ref{tab:training_hyperparameters}. We train for around 2K steps and validate every 500 steps to pick the best checkpoint for evaluation. For the baselines except TDPO and TI-DPO, we take the existing checkpoints from \citet{simpo2024} and use their hyperparameters.

For TDPO and TI-DPO, since the original method is applied on different task and training datasets, we retrain new checkpoints on our datasets. For TDPO, we use the TDPO$_2$ variant due to its superior performance over TPO$_1$, as reported in the original paper. We tune $\beta$ in the range $\{\text{1e-2}, \text{1e-1}\}$ and adopt the default for other hyperparameters. The final hyperparameters are $\beta=\text{1e-2}, \alpha=0.5, \text{Learning Rate}=\text{5e-6}$. For TI-DPO, we try different mixed-precision training setups. The original TIDPO implementation uses float32 for the main model and float16 for the reference model. We found that it worked much worse than our current setup, so we experiment with fp16 and bf16 and finally discover that bf16 works the best. We also tune $\beta$ in the range $\{\text{1e-2}, \text{1e-1}\}$ and Learning Rate in the range $\{\text{1e-6},\text{5e-6},\text{1e-5}\}$. We adopt the default for the other hyperparameters. The final hyperparameters are $\beta=\text{1e-2}, \alpha=0.5, \alpha_{\text{triplet}}=\text{1e-3}, \gamma=0.1, \lambda=0.7, \text{Learning Rate}=\text{1e-6}$.

There could be multiple reasons for the suboptimal results of TDPO and TIDPO in our setting.
% One reason could be the limited generalizability of TDPO and TIDPO to different model families and tasks. TDPO is originally evaluated on older models, including GPT-2-Large and Pythia 2.8B, and TIDPO is originally evaluated on verifiable tasks (MMLU, GSM8K, etc.).
Qualitatively, we inspect and compare their outputs with DPO’s on the ArenaHard benchmark, for LLaMA-3-8B-Instruct. For TDPO, we find that the model’s outputs are more hallucinatory than DPO’s on math and coding questions, and its outputs are more generic than DPO’s on technical writing questions. For TIDPO, across different types of queries, the model tends to produce outputs that are simpler and less comprehensive than DPO’s, and only addresses the core part of the user’s queries, despite having longer responses. For instance, for coding questions, TDPO’s programs have fewer explanatory comments. The fundamental causes of these behavioral differences need to be investigated further.

For OTPO, the authors use $\beta=\text{1e-2}, \text{Learning Rate}=\text{5e-7}$ for their training of LLaMA-3-8B-Instruct on UltraFeedback. We adopt their hyperparameters for our experiments.

\begin{table}[htp!]
    \centering
    \begin{tabular}{lc}
        \toprule
        Parameter & Value \\
        \midrule
         Max New Tokens & 4096 \\
         Temperature & 0.7 \\
         Top K & 50 \\
         Top P & 0.9 \\
         Presence Penalty & 0.1 \\
         Frequency Penalty & 0.1 \\
         Judge Max New Tokens & 1 \\
         Judge Temperature & 1.0 \\
         Baseline Model & GPT4-1106-preview \\
         \bottomrule
    \end{tabular}
    \caption{Hyperparameters for AlpacaEval}
    \label{tab:alpacaeval_hyperparameters}
\end{table}

\begin{table}[htp!]
    \centering
    \begin{tabular}{lc}
        \toprule
        Parameter & Value \\
        \midrule
         Max New Tokens & 1024 \\
         Temperature & varies \\
         Judge Max New Tokens & 2048 \\
         Judge Temperature & 0 \\
         \bottomrule
    \end{tabular}
    \caption{Hyperparameters for MT-Bench. We use the default temperature, which varies between different task subsets.}
    \label{tab:mtbench_hyperparameters}
\end{table}

\begin{table}[htp!]
    \centering
    \begin{tabular}{lc}
        \toprule
        Parameter & Value \\
        \midrule
         Max New Tokens & 2048 \\
         Temperature & 0 \\
         Judge Max New Tokens & 16000 \\
         Judge Temperature & 0 \\
         Baseline Model & GPT-4-0314 \\
         \bottomrule
    \end{tabular}
    \caption{Hyperparameters for ArenaHard}
    \label{tab:arenahard_hyperparameters}
\end{table}

\paragraph{Evaluation.} The default hyperparameters are used whenever they are available. See the details in Figure~\ref{tab:alpacaeval_hyperparameters},~\ref{tab:mtbench_hyperparameters}, and ~\ref{tab:arenahard_hyperparameters}.

\section{Additional Results}
\label{app:res}

\subsection{Multiple Runs}

We run three rounds of independent evaluations for all the checkpoints and report the mean and 95\% confidence interval in Table~\ref{tab:main_results_multiple_runs}.

\begin{table*}[]
    \small
    \centering
    \resizebox{\linewidth}{!}{
    \begin{tabular}{lcccccccccc}
        \toprule
         & \multicolumn{5}{c}{$\pi_{ref}$=LLaMA-3-8B-Base-SFT} & \multicolumn{5}{c}{$\pi_{ref}$=LLaMA-3-8B-Instruct} \\
         \cmidrule(lr){2-6} \cmidrule(lr){7-11}
         & \multicolumn{2}{c}{AlpacaEval} & MT & ArenaHard & \multirow{2}{*}{\textbf{Avg Rank $\downarrow$}} & \multicolumn{2}{c}{AlpacaEval} & MT & ArenaHard & \multirow{2}{*}{\textbf{Avg Rank $\downarrow$}} \\
         \cmidrule(lr){2-3} \cmidrule(lr){4-4} \cmidrule(lr){5-5} \cmidrule(lr){7-8} \cmidrule(lr){9-9} \cmidrule(lr){10-10}
         & LR $\uparrow$ & WR $\uparrow$ & Score $\uparrow$ & WR $\uparrow$ & & LR $\uparrow$ & WR $\uparrow$ & Score $\uparrow$ & WR $\uparrow$ & \\
         \midrule
         Base & 8.61$_{\pm 1.94}$ & 4.15$_{\pm 0.89}$ & 5.49$_{\pm 0.05}$ & 9.72$_{\pm 0.56}$ & 13.50 & 33.86$_{\pm 1.09}$ & 33.22$_{\pm 1.24}$ & 6.76$_{\pm 0.02}$ & 36.88$_{\pm 0.80}$ & 12.50 \\
         +RRHF & 11.68$_{\pm 0.60}$ & 6.75$_{\pm 0.26}$ & 6.18$_{\pm 0.05}$ & 25.71$_{\pm 0.40}$ & 9.00 & 44.31$_{\pm 1.18}$ & 37.92$_{\pm 0.55}$ & 6.67$_{\pm 0.06}$ & 41.80$_{\pm 0.61}$ & 9.63 \\
         +SLiC-HF & 16.83$_{\pm 0.35}$ & 11.98$_{\pm 0.24}$ & 5.93$_{\pm 0.03}$ & 13.31$_{\pm 1.12}$ & 10.25 & 38.75$_{\pm 1.26}$ & 37.45$_{\pm 0.83}$ & 6.77$_{\pm 0.02}$ & 42.73$_{\pm 0.82}$ & 9.38 \\
         +DPO & 21.90$_{\pm 1.57}$ & 17.02$_{\pm 0.91}$ & \textbf{6.58}$_{\pm 0.05}$ & 38.31$_{\pm 0.40}$ & 3.50 & 49.02$_{\pm 0.00}$ & 44.98$_{\pm 0.29}$ & 6.72$_{\pm 0.01}$ & \underline{55.32}$_{\pm 1.58}$ & 5.75 \\
         +TDPO & 10.67$_{\pm 1.44}$ & 5.49$_{\pm 0.76}$ & 5.71$_{\pm 0.03}$ & 13.24$_{\pm 1.14}$ & 11.88 & 36.98$_{\pm 1.03}$ & 36.74$_{\pm 0.99}$ & 6.63$_{\pm 0.02}$ & 39.35$_{\pm 1.61}$ & 12.00 \\
         +TIDPO & 8.33$_{\pm 1.74}$ & 4.23$_{\pm 0.70}$ & 5.32$_{\pm 0.03}$ & 9.79$_{\pm 0.37}$ & 13.50 & 34.98$_{\pm 0.45}$ & 34.10$_{\pm 0.50}$ & 6.78$_{\pm 0.02}$ & 37.51$_{\pm 1.21}$ & 11.00 \\
         +OTPO & \underline{22.44}$_{\pm 1.79}$ & 13.58$_{\pm 1.06}$ & 6.22$_{\pm 0.00}$ & 45.57$_{\pm 0.94}$ & 4.50 & 52.26$_{\pm 1.10}$ & 44.95$_{\pm 0.51}$ & 6.89$_{\pm 0.02}$ & 52.49$_{\pm 0.11}$ & 4.25 \\
         +IPO & 19.87$_{\pm 1.18}$ & 19.42$_{\pm 0.91}$ & 6.27$_{\pm 0.05}$ & 33.65$_{\pm 0.89}$ & 5.00 & 49.12$_{\pm 0.32}$ & 47.00$_{\pm 0.85}$ & \underline{6.98}$_{\pm 0.02}$ & 51.80$_{\pm 1.65}$ & \underline{4.00} \\
         +CPO & 19.14$_{\pm 19.14}$ & \underline{24.40}$_{\pm 0.31}$ & 6.13$_{\pm 0.00}$ & 36.82$_{\pm 0.92}$ & 5.25 & 42.04$_{\pm 0.78}$ & 45.80$_{\pm 0.76}$ & 6.67$_{\pm 0.01}$ & 50.59$_{\pm 1.75}$ & 7.88 \\
         +KTO & 18.92$_{\pm 18.92}$ & 15.58$_{\pm 0.70}$ & 5.95$_{\pm 0.02}$ & 17.61$_{\pm 0.68}$ & 8.25 & 43.66$_{\pm 1.70}$ & 40.15$_{\pm 1.41}$ & 6.77$_{\pm 0.02}$ & 47.01$_{\pm 1.95}$ & 7.63 \\
         +ORPO & 18.08$_{\pm 1.41}$ & 12.79$_{\pm 0.80}$ & 5.71$_{\pm 0.04}$ & 17.04$_{\pm 0.48}$ & 9.88 & 40.49$_{\pm 0.94}$ & 36.48$_{\pm 0.83}$ & 6.69$_{\pm 0.07}$ & 44.07$_{\pm 1.24}$ & 10.25 \\
         +RDPO & 22.23$_{\pm 1.98}$ & 15.94$_{\pm 1.11}$ & 6.32$_{\pm 0.04}$ & 28.12$_{\pm 0.79}$ & 4.75 & 50.48$_{\pm 1.69}$ & 45.61$_{\pm 1.37}$ & 6.84$_{\pm 0.05}$ & 54.18$_{\pm 1.15}$ & \underline{4.00} \\
         +SimPO & \textbf{29.78}$_{\pm 2.54}$ & \textbf{27.23}$_{\pm 1.99}$ & 6.10$_{\pm 0.04}$ & \underline{46.03}$_{\pm 1.32}$ & \underline{3.00} & \underline{57.00}$_{\pm 0.45}$ & \underline{48.16}$_{\pm 0.46}$ & 6.52$_{\pm 0.02}$ & \textbf{55.86}$_{\pm 1.06}$ & 4.75 \\
         +AttentionPO & 21.74$_{\pm 0.36}$ & 20.53$_{\pm 0.41}$ & \underline{6.54}$_{\pm 0.02}$ & \textbf{50.79}$_{\pm 1.86}$ & \textbf{2.75} & \textbf{60.20}$_{\pm 0.57}$ & \textbf{57.68}$_{\pm 0.60}$ & \textbf{7.14}$_{\pm 0.09}$ & 52.30$_{\pm 0.70}$ & \textbf{2.00} \\
         \bottomrule
    \end{tabular}
    }
    \caption{Main results from three rounds of independent evaluations on the same checkpoints. We report the mean and 95\% Confidence Interval across the evaluation runs.}
    \label{tab:main_results_multiple_runs}
    \vspace{-10pt}
\end{table*}

\subsection{Results on other models}

We additionally experiment with  zephyr-7b-sft-full \cite{tunstall2023}, which is a model supervised-finetuned from Mistral-7B-v0.1 \cite{mistral2023} on UltraChat-200K \cite{ultrachat2023}. We denote this model as \textbf{Mistral-7B-Base-SFT}. We use the same UltraFeedback binarized dataset \cite{ultrafeedback2024} (HuggingFaceH4/ultrafeedback\_binarized) and the same hyperparameters to train it. The results are shown in Table~\ref{tab:res_mistral_7b_base}.

\begin{table}[]
    \small
    \centering
    \begin{tabular}{lcccc}
        \toprule
         & \multicolumn{4}{c}{$\pi_{ref}$=Mistral-7B-Base-SFT} \\
         & \multicolumn{2}{c}{AlpacaEval} & MT & ArenaHard \\
         \cmidrule(lr){2-3} \cmidrule(lr){4-4} \cmidrule(lr){5-5}
         & LR $\uparrow$ & WR $\uparrow$ & Score $\uparrow$ & WR $\uparrow$ \\
         \midrule
         Base       & 8.77 & 4.20 & 4.54 & 2.26 \\
         +DPO & \underline{18.56}	& \underline{14.62} & \textbf{5.85}  & \underline{12.45} \\
         +\ourmethod & \textbf{22.39} & \textbf{19.47} & \underline{5.36} & \textbf{13.88} \\
         \bottomrule
    \end{tabular}
    \caption{Results on Mistral-7B-Base-SFT.}
    \label{tab:res_mistral_7b_base}
    \vspace{-5pt}
\end{table}

Compared with $\pi_{ref}$, \ourmethod~significantly enhances performances across different benchmarks. We observe a 14\% increase on AlpacaEval, a 0.82 increase on MT-Bench, and a 12\% increase on ArenaHard. In addition, \ourmethod~outperforms DPO in terms of AlpacaEval and ArenaHard, but DPO has a higher MT-Bench score.

\subsection{Using attentions from other models}
\label{app:res_attn_from_other_models}

\begin{table*}[]
    \small
    \centering
    \begin{tabular}{lcccccccc}
        \toprule
         & \multicolumn{4}{c}{$\pi_{ref}$=LLaMA-3-8B-Base-SFT} & \multicolumn{4}{c}{$\pi_{ref}$=LLaMA-3-8B-Instruct} \\
         & \multicolumn{2}{c}{AlpacaEval} & MT & ArenaHard & \multicolumn{2}{c}{AlpacaEval} & MT & ArenaHard \\
         \cmidrule(lr){2-3} \cmidrule(lr){4-4} \cmidrule(lr){5-5} \cmidrule(lr){6-7} \cmidrule(lr){8-8} \cmidrule(lr){9-9} 
         & LR $\uparrow$ & WR $\uparrow$ & Score $\uparrow$ & WR $\uparrow$ & LR $\uparrow$ & WR $\uparrow$ & Score $\uparrow$ & WR $\uparrow$ \\
         \midrule
         Base       & 9.20 & 4.63 & 5.47 & 9.92 & 33.60 & 32.60 & 6.78 & 37.57 \\
         +DPO       & \underline{22.88} & 17.70 & \textbf{6.54} & 38.54 & \underline{49.02} & \underline{44.72} & 6.73 & \textbf{53.98} \\
         +\ourmethod & 21.55 & \underline{20.23} & \underline{6.52} & \underline{49.72} & \textbf{60.77} & \textbf{58.29} & \textbf{7.19} & \underline{52.06} \\
         w/ attn from sft & - & - & - & - & 48.54 & 44.67 & \underline{6.99} & 45.71 \\
         w/ attn from inst & \textbf{26.97} & \textbf{27.74} & 6.44 & \textbf{58.21} & - & - & - & - \\
         \bottomrule
    \end{tabular}
    \caption{Results of using attentions from other models.}
    \label{tab:res_attn_from_other_models}
    \vspace{-5pt}
\end{table*}

We investigate whether attention weights from one model can be used to train another model. In particular, we use the attentions from LLaMA-3-8B-Base-SFT to train LLaMA-3-8B-Instruct (\texttt{w/ attn from sft}) and vice versa (\texttt{w/ attn from inst}). The results are shown in Table~\ref{tab:res_attn_from_other_models}.

Training the SFT model with attentions from the stronger Instruct model (\texttt{w/ attn from inst}) leads to a further increase in performance on AlpacaEval and ArenaHard on top of the original \ourmethod~results. In contrast, training the instruction model with attentions from the SFT model (\texttt{w/ attn from sft}) gives worse performance than the original \ourmethod, but the results are nonetheless better than $\pi_{ref}$ by a significant margin across benchmarks. This suggests that attentions from a stronger model gives stronger performance. \ourmethod~can still deliver performance gains even with attentions from a model weaker than $\pi_{ref}$, but to a lesser extent.

\subsection{Analysis of Response Length}
\label{app:resp_len}

\begin{table}[H]
    \small
    \centering
    \begin{tabular}{lcccc}
        \toprule
         & \multicolumn{2}{c}{$\pi_{ref}$=SFT} & \multicolumn{2}{c}{$\pi_{ref}$=Instruct} \\
         \cmidrule(lr){2-3} \cmidrule(lr){4-5}
         & WR $\uparrow$ & \#Tokens & WR $\uparrow$ & \#Tokens \\
         \midrule
         Base             & 9.92  & 1254 & 37.57 & 580 \\
         +RRHF            & 26.00 & \underline{1506} & 41.91 & 506 \\
         +SLiC-HF         & 13.08 & 556 & 42.76 & 575 \\
         +DPO             & 38.54 & 921 & 53.98 & 533 \\
         +TDPO            & 12.09 & 1233 & 41.00 & 565 \\
         +TI-DPO          & 10.14 & 1238 & 38.73 & \underline{585} \\
         +IPO             & 34.56 & 675 & 50.50 & 546 \\
         +CPO             & 36.04 & \textbf{1588} & 52.07 & \textbf{631} \\
         +KTO             & 18.28 & 516 & 48.98 & 535 \\
         +ORPO            & 17.32 & 594 & 44.64 & 540 \\
         +R-DPO           & 28.72 & 731 & \textbf{54.98} & 524 \\
         +SimPO           & \underline{47.01} & 796 & \underline{54.94} & 505 \\
         +\ourmethod      & \textbf{49.72} & 1157 & 52.06 & 553 \\
         \bottomrule
    \end{tabular}
    \caption{ArenaHard Results. WR: Win Rate against GPT-4-0314. \#Tokens: Number of response tokens.}
    \label{tab:resp_len}
    \vspace{-5pt}
\end{table}

We report the length of generated responses on ArenaHard in Table~\ref{tab:resp_len}. For both the SFT and Instruct model, responses produced by \ourmethod~are not longer than those from either the base model or other baseline methods. This shows that the performance boost does not come from verbosity.

\section{Examples}
\label{app:examples}

\begin{figure*}
    \centering
    \includegraphics[width=\linewidth]{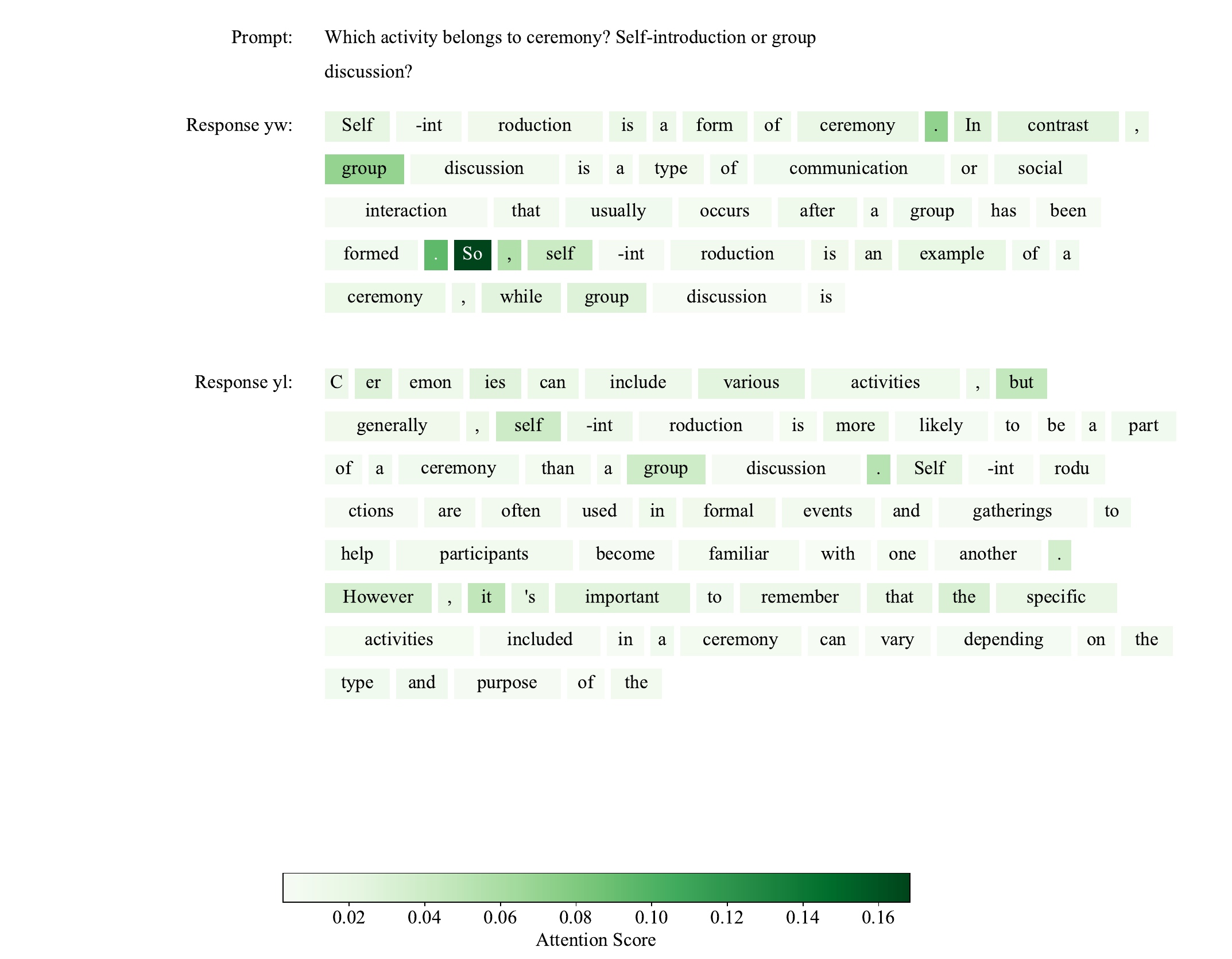}
    \caption{An example for LLaMA-3-8B-Base-SFT.}
    \label{fig:example_sft}
\end{figure*}

\begin{figure*}
    \centering
    \includegraphics[width=\linewidth]{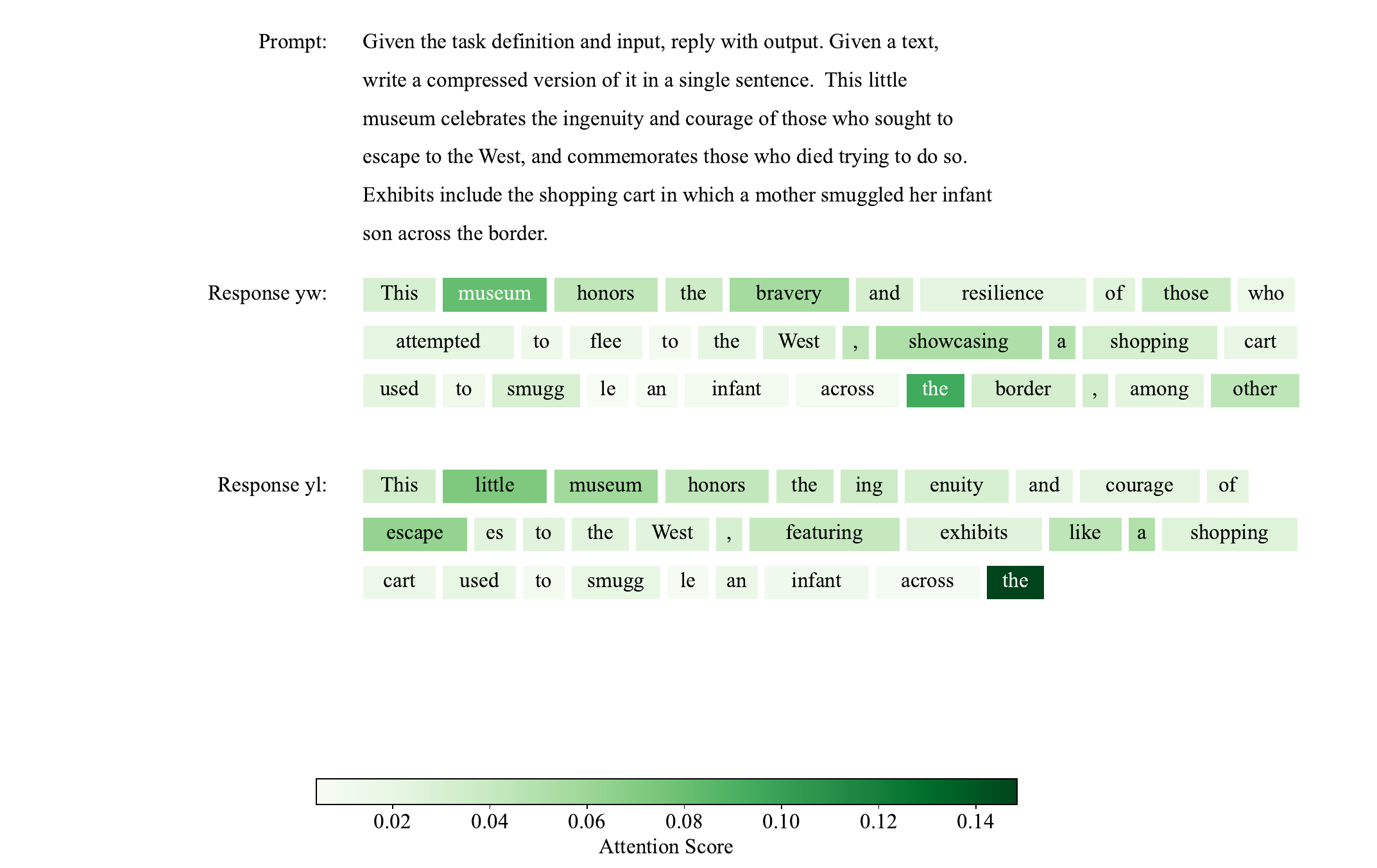}
    \caption{An example for LLaMA-3-8B-Instruct.}
    \label{fig:example_inst}
\end{figure*}

We show an example of attention weights for LLaMA-3-8B-Base-SFT in Figure~\ref{fig:example_sft} and LLaMA-3-8B-Instruct in Figure~\ref{fig:example_inst}.

\section{Compute}

We run all experiments on a single node with 40 CPU cores, 256 GB of memory, and 2 Nvidia A100 80 GB GPUs.

\section{Prompt}
\label{app:prompt}

\begin{figure*}
    \centering
    \includegraphics[width=\linewidth]{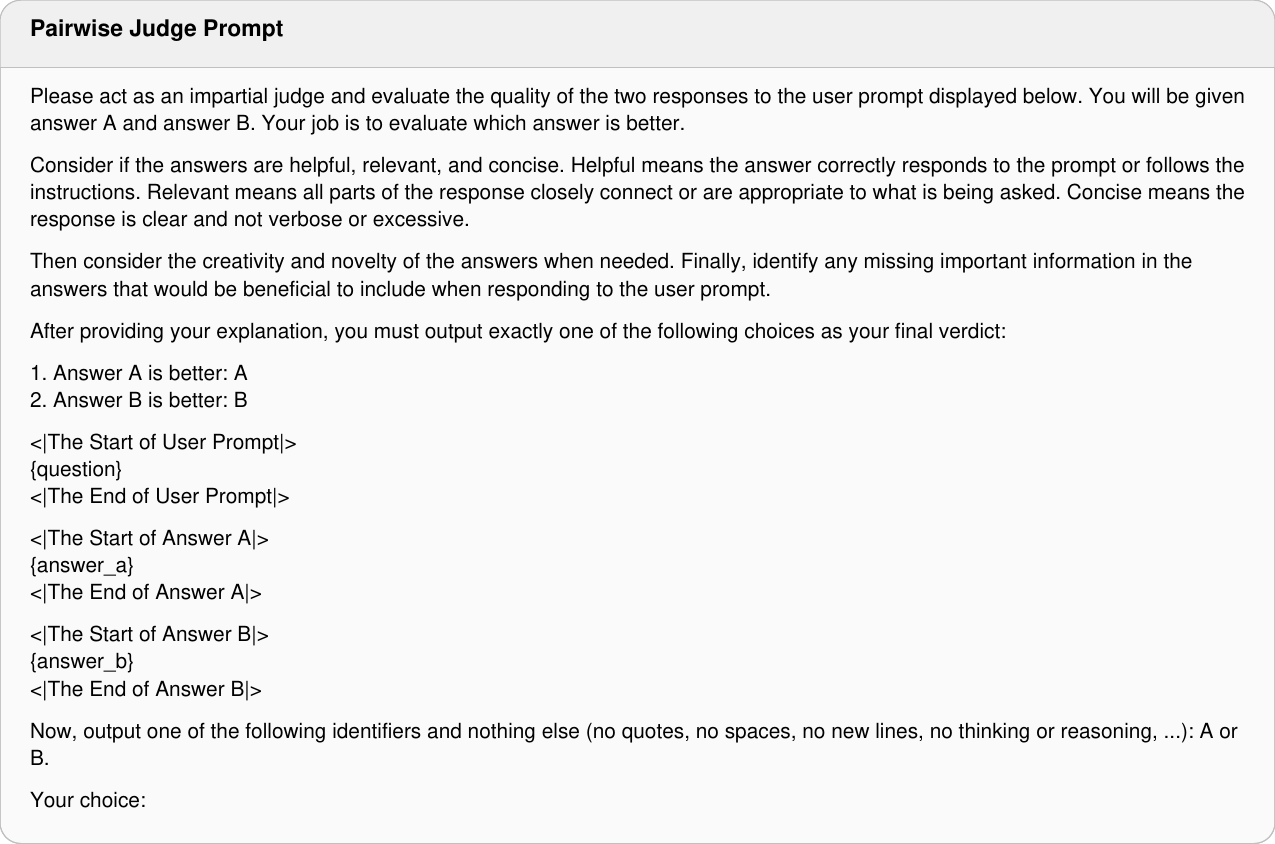}
    \caption{Pairwise Judge Prompt}
    \label{fig:pairwise_judge_prompt}
\end{figure*}

\begin{figure*}
    \centering
    \includegraphics[width=\linewidth]{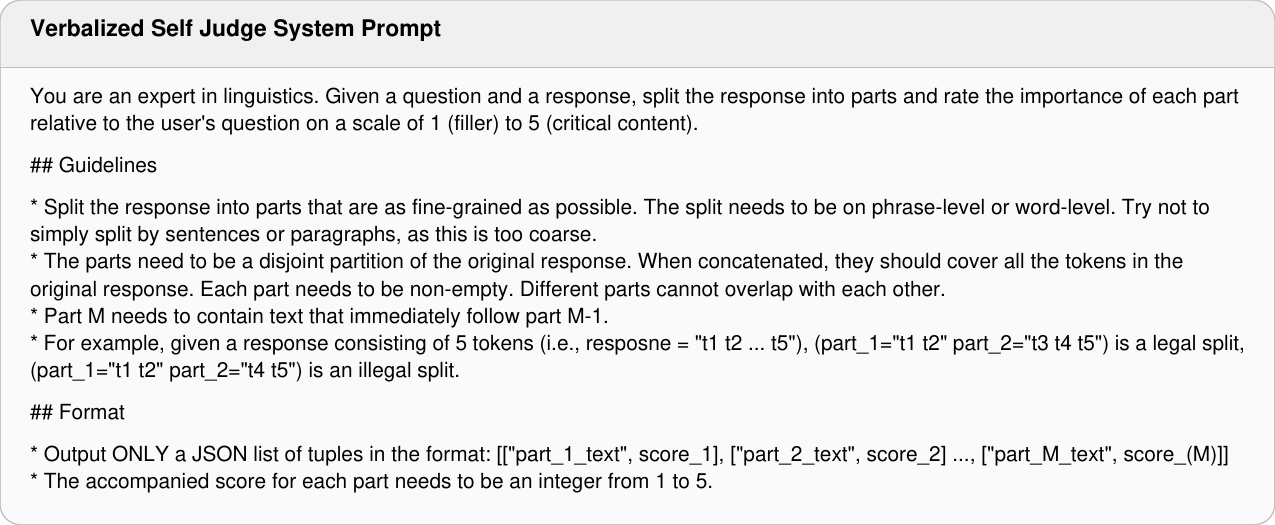}
    \caption{Verbalized Self-judge System Prompt}
    \label{fig:verbalized_self_judge_system_prompt}
\end{figure*}

\begin{figure*}
    \centering
    \includegraphics[width=\linewidth]{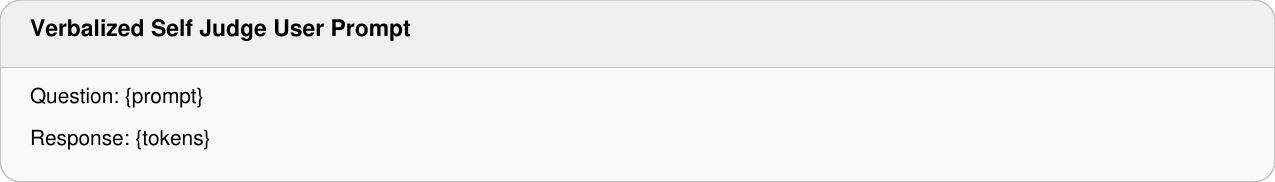}
    \caption{Verbalized Self-judge User Prompt}
    \label{fig:verbalized_self_judge_user_prompt}
\end{figure*}

\end{document}